\documentclass[a4paper,fleqn]{cas-sc}

\usepackage[numbers]{natbib}
\usepackage{bbm}
\usepackage{bm}
\usepackage{amsfonts}
\usepackage{amssymb}
\usepackage{multirow}
\usepackage{caption}
\usepackage{subcaption}
\usepackage[flushleft]{threeparttable}
\usepackage{newtxtext}
\usepackage{setspace}
\usepackage[utf8]{inputenc}
\usepackage{textcomp}
\setbool{casreviewlayout}{true}

\def\tsc#1{\csdef{#1}{\textsc{\lowercase{#1}}\xspace}}
\tsc{WGM}
\tsc{QE}

\begin{document}
\let\WriteBookmarks\relax
\def\floatpagepagefraction{1}
\def\textpagefraction{.001}

\shorttitle{Problem dependent attention and effort}    

\shortauthors{C. Rohlfs}  

\title [mode = title]{Problem-dependent attention and effort in neural networks with applications to image resolution and model selection}  



%

\author{Chris Rohlfs}[orcid=0000-0001-7714-9231]



\ead{car2228@columbia.edu}



\affiliation{organization={Columbia University Department of Electrical Engineering},
            addressline={Mudd 1310, 500 West $120^{th}$ Street}, 
            city={New York},
            postcode={10027-6623}, 
            state={NY},
            country={USA}}


\begin{abstract}
This paper introduces two new ensemble-based methods to reduce the data and computation costs of image classification. They can be used with any set of classifiers and do not require additional training. In the first approach, data usage is reduced by only analyzing a full-sized image if the model has low confidence in classifying a low-resolution pixelated version. When applied on the best performing classifiers considered here, data usage is reduced by 61.2\% on MNIST, 69.6\% on KMNIST, 56.3\% on FashionMNIST, 84.6\% on SVHN, 40.6\% on ImageNet, and 27.6\% on ImageNet-V2, all with a less than 5\% reduction in accuracy. However, for CIFAR-10, the pixelated data are not particularly informative, and the ensemble approach increases data usage while reducing accuracy. In the second approach, compute costs are reduced by only using a complex model if a simpler model has low confidence in its classification. Computation cost is reduced by 82.1\% on MNIST, 47.6\% on KMNIST, 72.3\% on FashionMNIST, 86.9\% on SVHN, 89.2\% on ImageNet, and 81.5\% on ImageNet-V2, all with a less than 5\% reduction in accuracy; for CIFAR-10 the corresponding improvements are smaller at 13.5\%. When cost is not an object, choosing the projection from the most confident model for each observation increases validation accuracy to 81.0\% from 79.3\% for ImageNet and to 69.4\% from 67.5\% for ImageNet-V2.\end{abstract}


\maketitle

\section{Introduction} \label{downsampling introduction}

Research effort should increase with the importance and difficulty of the problem. In biological systems, learning is typically curiosity-driven, with new observations examined because they have some new features not captured by the existing framework \cite{Sinz2019}. By contrast, artificial neural networks typically apply a uniform process to all observations, ignoring the difficulty associated with specific test cases. This dominant approach is wasteful; for many observations, the correct classification can be accurately obtained with considerably less resource use. Computing costs limit the complexity of deep learning models and delay the development of new approaches, and they are rapidly rising. AlexNet \cite{Krizhevsky2014}, the winner of the ILSVRC 2012 competition, classifies an image in 6.8 ms; later influential models such as ResNeXt-101-32x8d \cite{Xie2017} and VGG-19-bn \cite{Simonyan2015} take 9.9 and 23 times longer, at 67.4 ms and 157 ms, respectively.

This study introduces an ensemble-based forecasting approach that only incurs large resource costs on hard-to-classify cases. For each observation, the simplest classifier is used first; the more complex classifier’s output is only generated for cases in which the simple model lacks confidence in its projection. Targeting the use of resource-intensive classifiers in this way substantially reduces memory and computing costs on average with minimal loss in accuracy.

One innovative line of research in dynamic deep learning performs preliminary analysis on coarse representations of images to identify Regions of Interest (RoI) for further analysis \cite{Cordonnier2021, Gao2018, Yang2020b, Yuan2019a}. Like the current study, these RoI-based approaches employ models' self-assessments of projection confidence, and they are specially trained to balance accuracy and computing cost for specific observations. Relative to that strategy, the ensemble method developed here has the advantage that it is versatile and easy to use. The ensemble can include any combination of deep and shallow classifiers; it employs existing trained models and does not require any additional fitting or parameter estimation.

Two applications of the proposed ensemble approach are evaluated. The first reduces data usage by initially classifying low-resolution pixelated versions of images and only analyzing the full-sized image when necessary. Strategies of this form are valuable when large numbers of cases must be classified in real time or when the data themselves are costly to acquire, as with the chest x-ray data examined by \cite{Sabottke2020} and the \$3,000 per subject brain MRI data by \cite{Spreng2022}. The approach is tested here using common datasets from the computer vision literature. When applied on the best performing network-based classifiers considered here, data usage is reduced by 61.2\% on MNIST, 69.6\% on KMNIST, 56.3\% on FashionMNIST, 84.6\% on SVHN, 40.6\% on ImageNet, and 27.6\% on ImageNet-V2, all with a less than 5\% reduction in accuracy. For the CIFAR-10 dataset, which requires classifying thumbnail images of objects and animals in varying contexts and positions, the categories are too complex to reliably distinguish from the low-resolution images, and using an ensemble increases data use and reduces accuracy. Performance thus varies in understandable ways---the ensemble is effective when the model truly does not need all the data that it consumes. 

The second application reduces computing time by initially classifying images using simpler classifiers with few parameters, only applying complex, time-intensive classifiers when necessary. Computation cost is reduced by 82.1\% on MNIST, 47.6\% on KMNIST, 72.3\% on FashionMNIST, 86.9\% on SVHN, 89.2\% on ImageNet, and 81.5\% on ImageNet-V2, all with a less than 5\% reduction in accuracy; for CIFAR-10 the corresponding improvements are smaller at 13.5\%. When cost is not an object, choosing the projection from the most confident model for each observation increases validation accuracy to 81.0\% from 79.3\% for ImageNet and to 69.4\% from 67.5\% for ImageNet-V2.

\section{Recent Literature} \label{downsampling recent literature}

The approach introduced in this study is \emph{ensemble-based} in that it combines the outputs from multiple classifiers. But unlike popular model averaging \cite{Hinton2015, Lee2015} or meta-learning \cite{Wolpert1992} ensemble approaches, which typically increase performance and resource use, the aim here is to \emph{reduce} resource-intensity while accepting slight reductions in accuracy. This paper builds most directly on prior work in two areas: the use of reduced-resolution images in deep learning networks and work to improve neural networks' efficiency.

\subsection{Image Resolution}

As super-resolution and de-noising approaches (\emph{cf.} \cite{Yang2019b, Wang2020}) illustrate, additional detail can often be accurately inferred from coarse images. Nevertheless, in typical classification settings, using lower-resolution images noticeably degrades neural networks' performance \cite{Koziarski2018}. Recent work improves recognition of low-resolution faces by training networks with matched low and high resolution examples \cite{Massoli2020} or with low-resolution images at different scales \cite{Mishra2021}. In both studies, the authors produce network-based classifiers that identify facial features that are robust across multiple levels of image resolution. The resulting classifiers learn from high-resolution data available in training but are optimized for application in low-resolution settings (such as surveillance) that are common in practice.

Like this previous research, this study uses images at different resolution levels; however, the setting considered is somewhat different. In the surveillance studies, only the low resolution data are available at the time the model is deployed. In the current setting, the researcher has access at the time of deployment to images at multiple resolution levels, but using the test data from the higher resolution cases entails some extraction or computation cost.

\subsection{Effort and Efficiency}

Much of the work to reduce the resource intensity of deep learning models has focused on simplifying the model structure, an area known as \emph{model compression} \cite{Han2016a, Han2016b}. Two similar approaches, quantization and pruning, reduce the amount of information that must be stored when models are trained: quantization rounds the connection weights so that they require less memory to store and process \cite{Gong2014, Jacob2018, Krishnamoorthi2018}, and pruning rounds smaller weights to zero (\emph{i.e.,} removes them \cite{Han2015, Li2017b}). Knowledge distillation approaches perform transfer learning, where a large, resource-intensive network serves as the ``teacher,'' and a simplified version learns to replicate its behavior \cite{Hinton2015, Polino2018}. The network architectures of many influential classifiers---including the MobileNet v3 \cite{Howard2019} and EfficientNet \cite{Tan2020} models used in this study---employ multiple such compression approaches to reduce computation costs.

In addition to these compression-based techniques that reduce resource-intensity for the model as a whole, some previous studies in the area of \emph{dynamic neural networks} do so observation-by-observation. These models do so through routing within the structure of the network---so that simpler problems receive less analysis (\cite{Han2022}); this routing is often operationalized through \emph{early exit}, whereby if a partial run of the model produces a sufficiently confident projection, then analysis stops and that projection is used. Exiting in this way helps to reduce the number of computationally-intensive convolution steps that are required for classifying easier cases. Exit rules depend in different cases on thresholds for confidence in the chosen classification \cite{Kaya2019, Kaya2019b, Figurnov2017}, the distribution of probabilities across classes \cite{Teerapittayanon2016}, or more complex rules involving the computational budget and cost of further analysis \cite{Bolukbasi2017, Huang2018}.

As noted in the introduction, one important area of deep learning research that employs dynamic routing of this form is in the analysis of complex visual inputs. Some researchers analyze such data by first passing over downsampled coarse representations of the images to identify Regions of Interest (RoIs)---and then later extracting and analyzing full resolution versions of those segments. \cite{Gao2018} use such an RoI-based approach to detect the presence of pedestrians in images of public locations such as crosswalks, and \cite{Cordonnier2021} employ a similar strategy to detect multiple objects in images. \cite{Yuan2019a} apply a sequential approach of this form to interpolate missing video frames. The authors first use coarse representations of the images to identify movement and RoIs, and they obtain refined projections based upon full-resolution segments pulled from those RoIs. In some cases, as in \cite{Yang2020b}, who focus on CIFAR-10, CIFAR-100, and ImageNet images, the number of RoIs vary and can be zero---implying early exit. Those authors obtain a result that aligns with the findings here, that the canonical representation of an object (\emph{e.g.,} a perched owl looking straight into the camera) can be correctly labeled using a simpler model, while variations on this representation (\emph{e.g.,} an owl captured from a different angle or pictured together with a person) require a more resource-intensive classification approach.

The studies described above provide valuable methods for balancing resource use and accuracy in image classification, but they require researchers to develop and train new networks for each application. By introducing an ensemble-based strategy that can be used with any set of trained classifiers, this study contributes a versatile tool that can be easily applied to a wide range of resource management problems.

\section{Conceptual Framework} \label{downsampling conceptual framework}

\begin{figure}
\centering
\begin{subfigure}{0.4\textwidth}
  \centering
  \includegraphics[width=\textwidth]{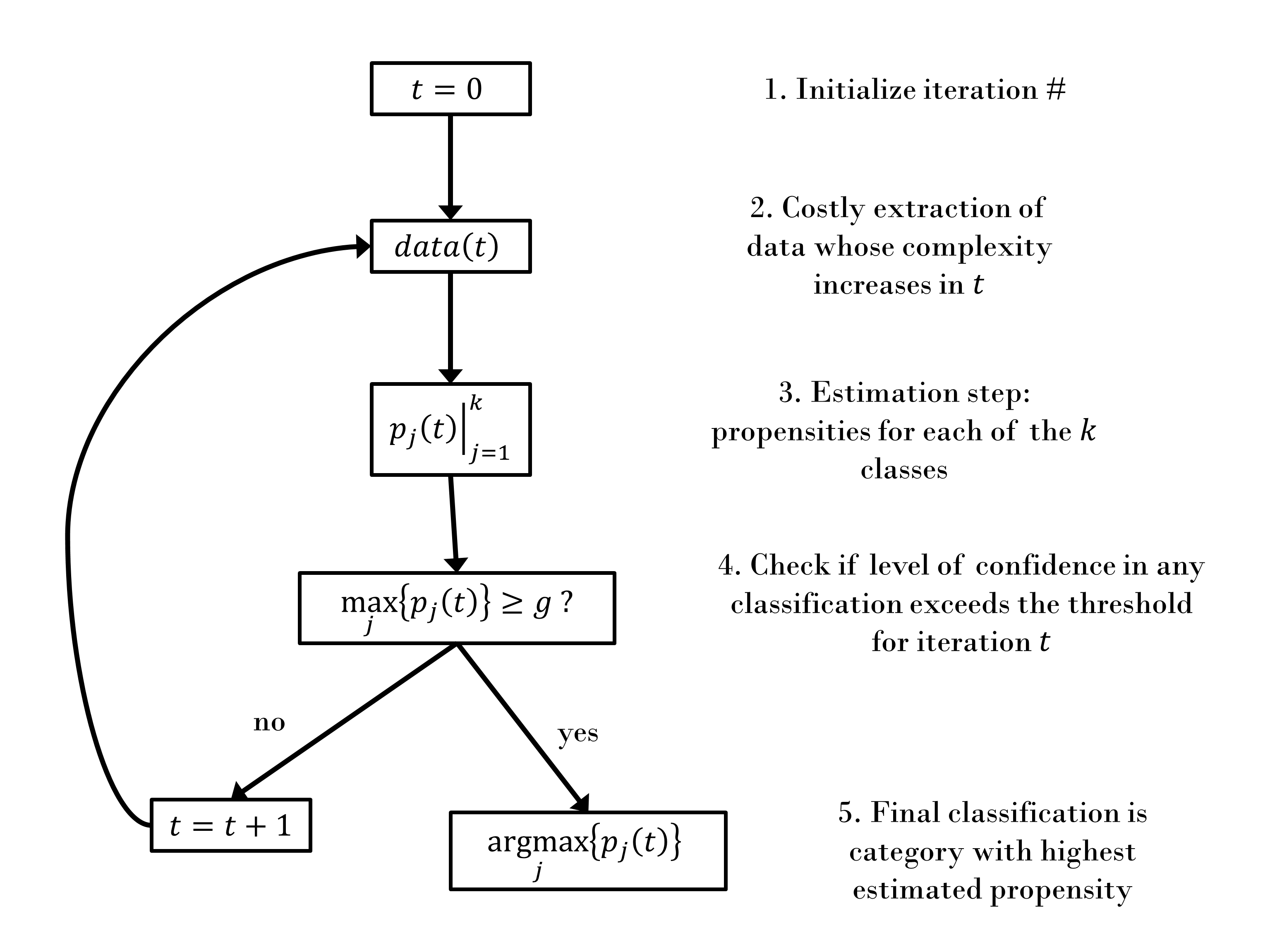}
  \caption{Image Resolution Application}
  \label{fig:flowchart1}
\end{subfigure}
\begin{subfigure}{0.4\textwidth}
  \centering
  \includegraphics[width=\textwidth]{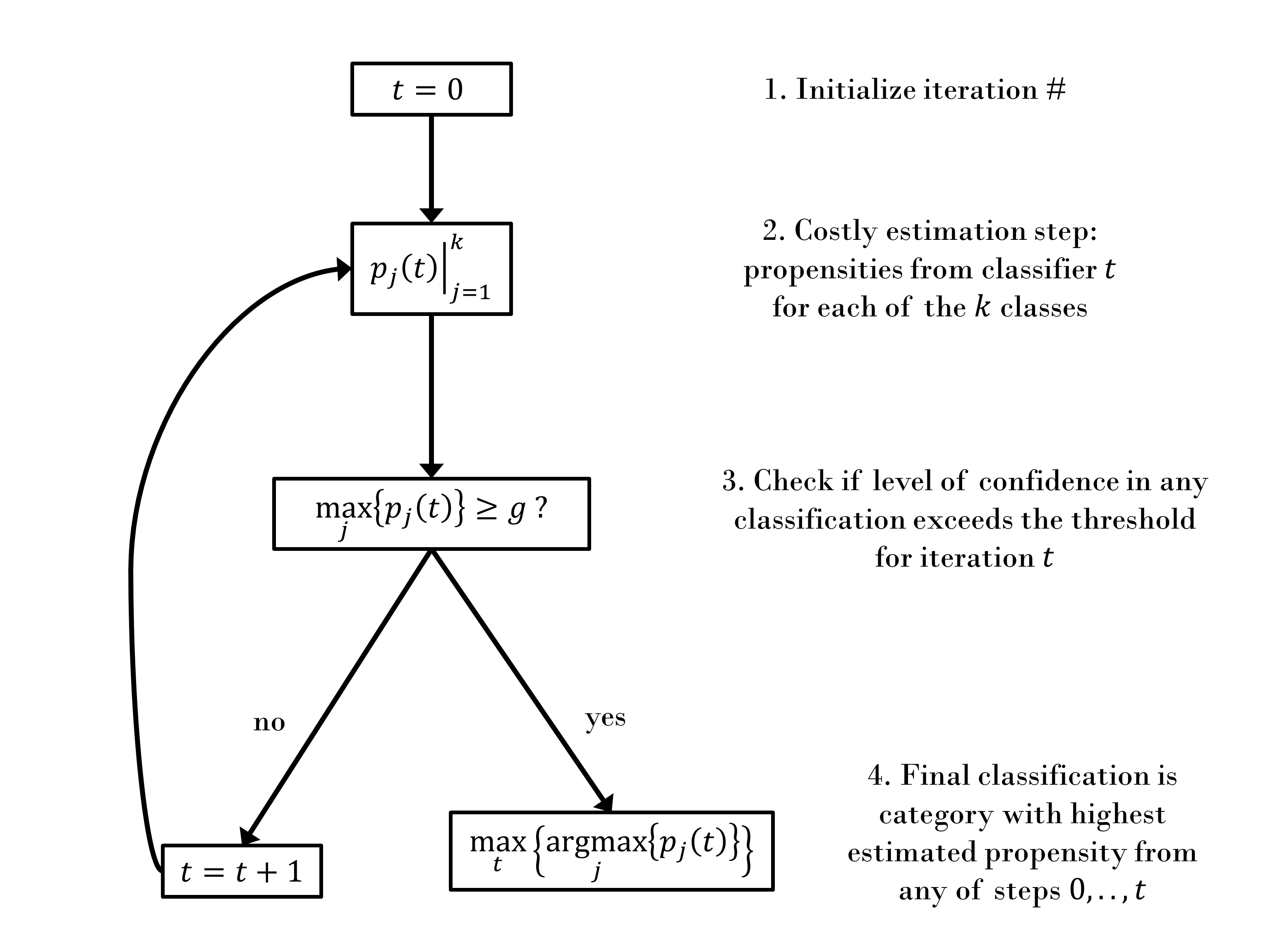}
  \caption{Computational Complexity Application}
  \label{fig:flowchart2}
\end{subfigure}
\caption{Flow Charts Illustrating Deployment of Problem-Dependent Effort Classifier on a Single Test Case}
\label{fig:flowchart}
\end{figure}

The researcher has access to an array of classification approaches with varying degrees of complexity and resource intensity, and any of those approaches can be applied for a given test case. The approaches are organized in increasing order of resource consumption and indexed by $t$. In the first application considered in this study, illustrated by Figure \ref{fig:flowchart1},  each of the classification approaches uses the same trained model, but in some cases, the input is a coarse representation of the image being classified, while in other cases, it is the full-sized image. Step 1 initializes the iteration number at $t=0$. The coarsest representation of the image is then extracted in step 2. A classifier is then used in step 3 to assign probabilities $p_1 . . . p_{k}$ to each of the possible classifications. For the purposes of this analysis, let these model-specific probabilities be denoted propensities. Next, step 4 determines how confident the model is in its classification, examining the distribution of these propensities across the $k$ categories. If the propensities are concentrated on one classification---in the extreme case 100\% for one category and 0\% for the others---then the classifier is highly confident in its selection. If the propensities are relatively spread out---in the extreme case with values of 10\% for each of 10 categories---then the model and data provide little information value, and the clasifier has little confidence in its selection. Step 4 compares the propensity score for the most promising candidate classification to some threshold. If the classifier is highly confident in its selection, then the classifier stops at stage 5 and chooses this category with the highest propensity score. If the classifier's level of confidence in its choice falls below the threshold, then the iterator $t$ increments by one, and the process repeats with a slightly higher resolution image.

In the second application, illustrated by Figure \ref{fig:flowchart2}, each of the classification approaches is a different trained model---where some of those models are more computationally intensive than others. Hence, the cost-intensive step is now the application of the model to estimate the $k$ propensities of each of the different classes being considered, which is performed in step 2. In the first application, these propensities vary across iterations $t$ because the input data change. In this second application, the propensities vary with $t$ because different models are used.

Another difference between the two classification approaches can be seen in the last step in the charts. In the first application, once the forecast using the higher resolution image is known, the projection using the low-resolution image can be discarded. Thus, the final classification is chosen by performing a single argmax over $k$ possibilities. For the second application, a given test case might have features that one of the less computationally intensive models is better suited to discern and classify. Consequently, at each stage, the projected value that is used is the one that has the highest propensity that has been encountered so far.

In order for the proposed approaches to be viable, two preconditions must be met. First, the models' internal propensity scores---\emph{i.e.,} their self-assessed levels of confidence in their selected classifications---must be informative. Deep learning models have been found in many cases to be overconfident in their projections, so that the classifications they predict with probability $p$ are correct less than $100*p\%$ of the time \cite{Nguyen2015, Gal2016}. For the purposes of this analysis, the self-assessments do not need to be unbiased; they simply must increase with the accuracy of the classification. As demonstrated in Section \ref{downsampling results}, the propensity scores used in this study generally meet this requirement, with the exception of the lowest resolution images that are excluded from the ensembles.

The second requirement is that there exist low-cost alternatives to the best or most costly classifier---and that those alternatives sometimes produce accurate classifications. This second condition naturally depends upon the types of costs and classifications being considered. The ensemble's effectiveness is limited to cases in which there is some inefficiency---a costly resource that is being expended by the model but is not critical to the classification. 

There is not a formal process for determining the threshold $g$; it depends upon the relative importance of resource efficiency and accuracy, which are application-specific and often not quantified. Supposing that the values $p_j(t)$ are unbiased or an appropriate bias correction is applied, then $g$ can be set to the minimum level of accuracy required of the final classifier. Additionally, if there is a holdout sample for which all the classifers' projections are available, the tradeoffs can be better understood by plotting average resource use and classification accuracy for different values of $g$. For the models and datasets considered in this study, plots of this form are presented in Figures \ref{fig:ppf} and \ref{fig:gmacs_ppf}.

\section{Data} \label{downsampling data}

\begin{figure*}
\centering
\begin{subfigure}{0.4\textwidth}
  \centering
  \includegraphics[width=\textwidth]{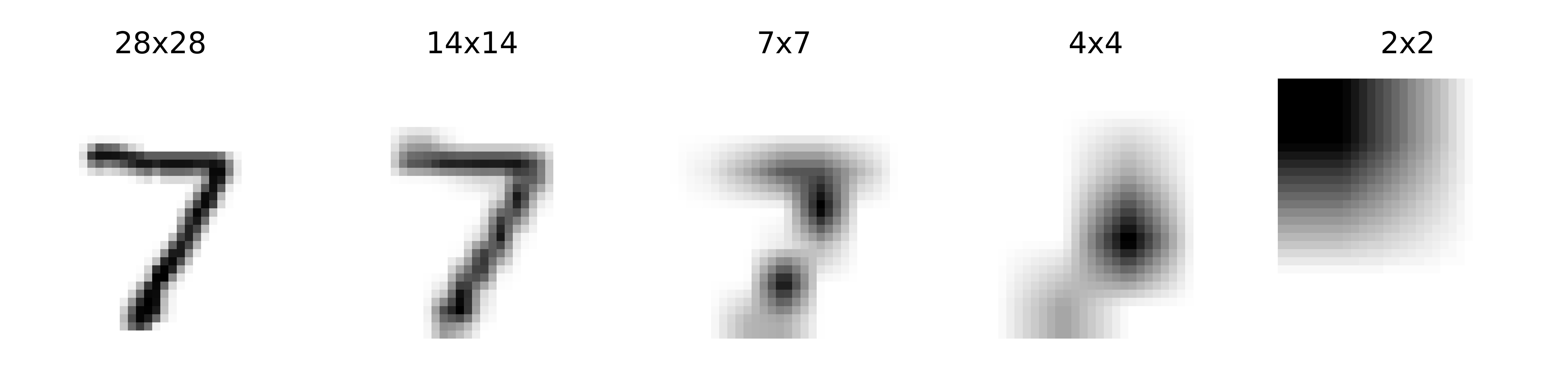}
  \caption{MNIST, Label = 7}
  \label{fig:mnist}
\end{subfigure}
\begin{subfigure}{0.4\textwidth}
  \centering
  \includegraphics[width=\textwidth]{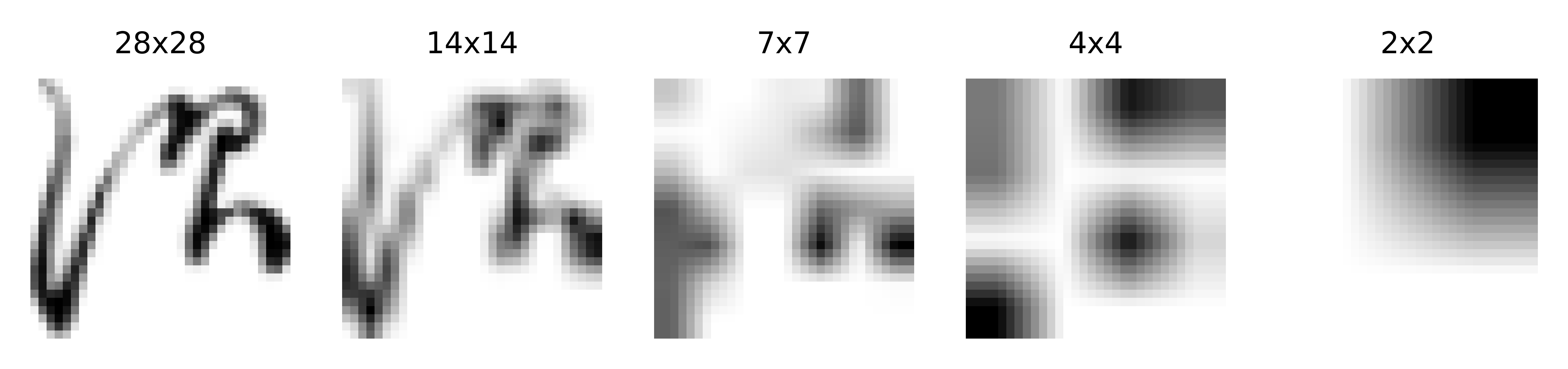}
  \caption{KMNIST, Label = ``su''}
  \label{fig:kmnist}
\end{subfigure}
\begin{subfigure}{0.4\textwidth}
  \centering
  \includegraphics[width=\textwidth]{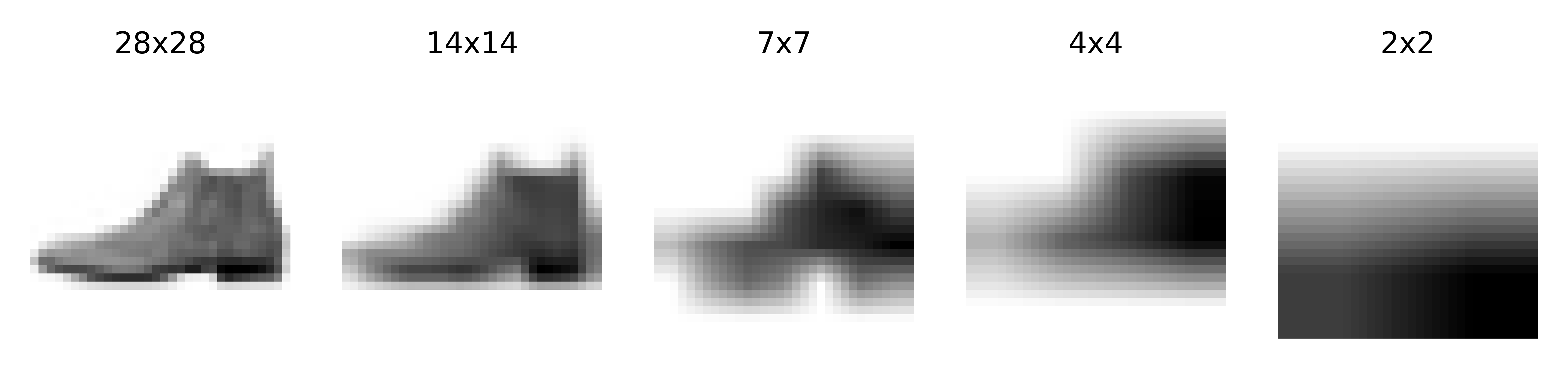}
  \caption{Fashion MNIST, Label = ``ankle boot''}
  \label{fig:fashionmnist}
\end{subfigure}
\begin{subfigure}{0.4\textwidth}
  \centering
  \includegraphics[width=\textwidth]{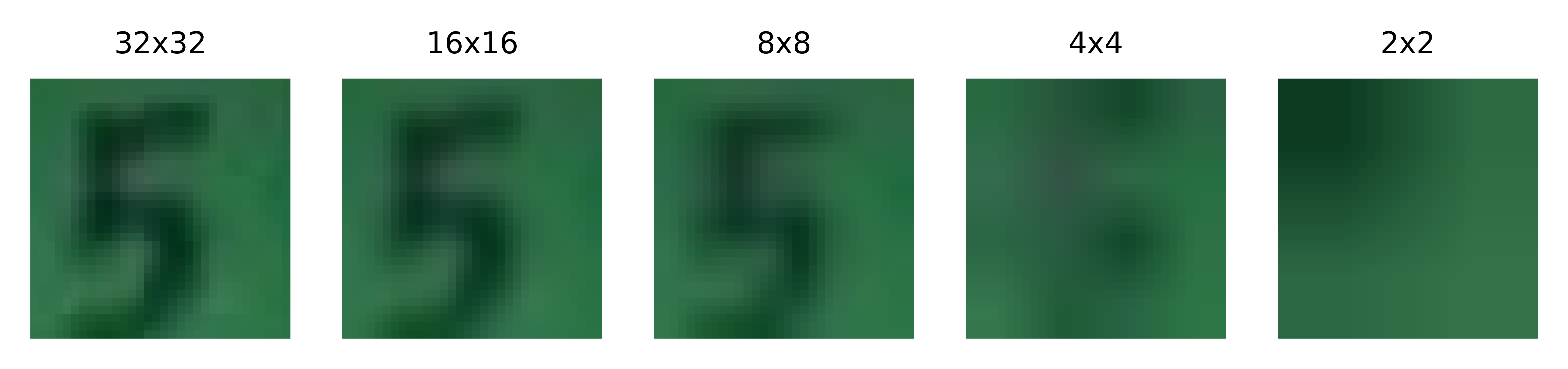}
  \caption{SVHN, Label = 5}
  \label{fig:svhn}
\end{subfigure}
\begin{subfigure}{0.4\textwidth}
  \centering
  \includegraphics[width=\textwidth]{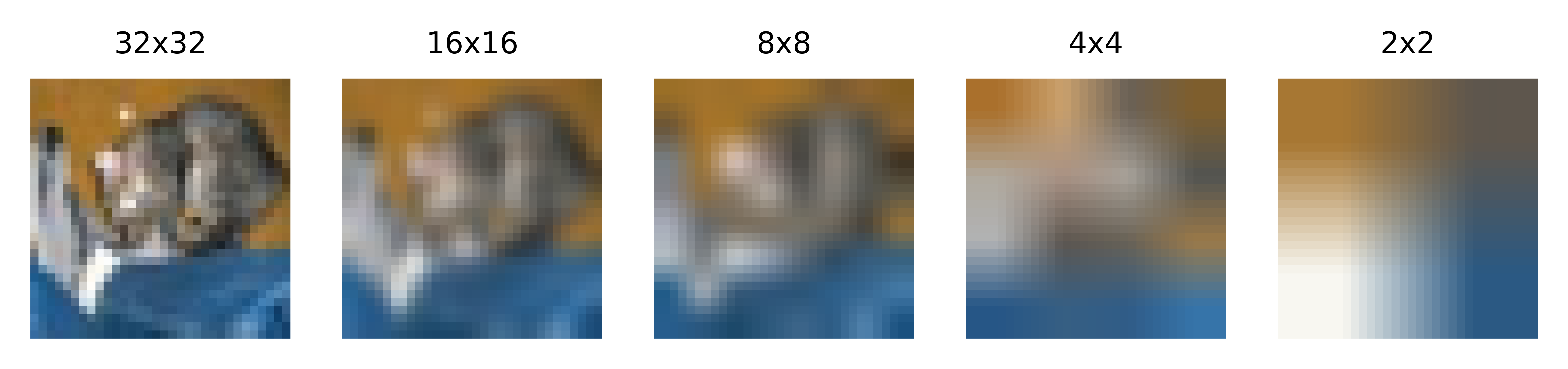}
  \caption{CIFAR-10, Label = ``cat''}
  \label{fig:cifar}
\end{subfigure}
\begin{subfigure}{0.4\textwidth}
  \centering
  \includegraphics[width=\textwidth]{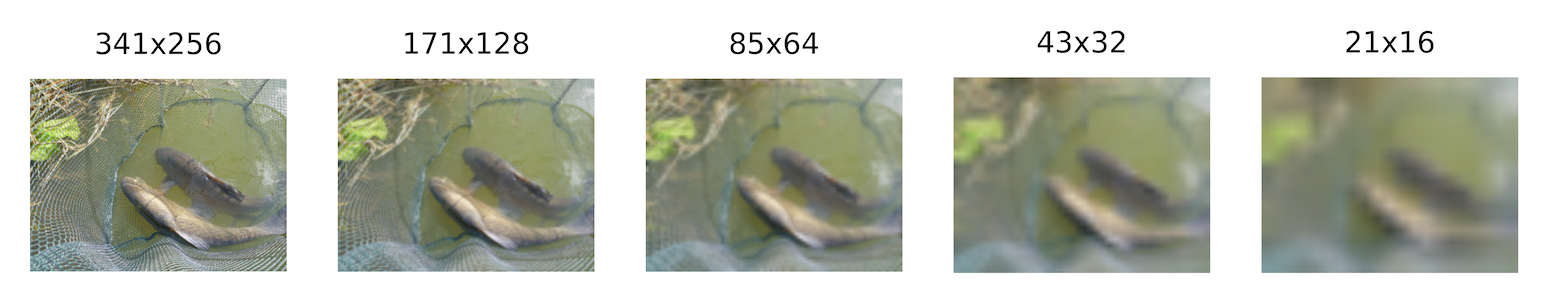}
  \caption{ImageNet, Label = ``tench''}
  \label{fig:imagenet}
\end{subfigure}
\begin{subfigure}{0.4\textwidth}
  \centering
  \includegraphics[width=\textwidth]{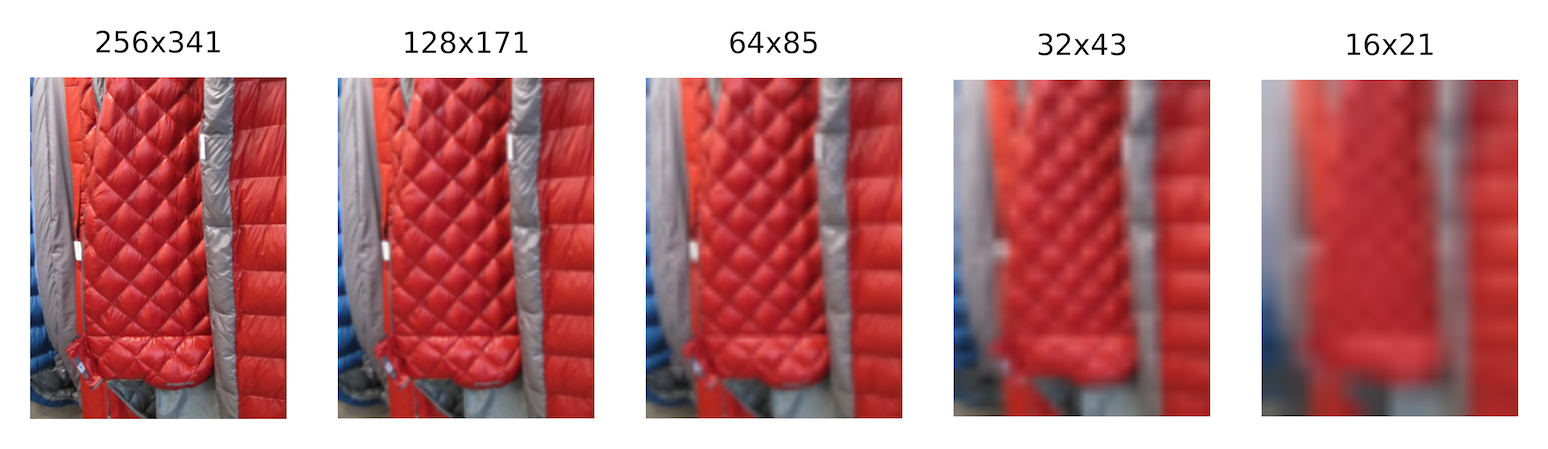}
  \caption{ImageNet-V2, Label = ``sleeping bag''}
  \label{fig:imagenet2}
\end{subfigure}
\begin{minipage}{0.7\linewidth}
\vspace{0.10 in}
\flushleft \small Notes: Within each of the seven panels, the first image shows the first observation from the test or validation sample. The caption gives the true label that is assigned to that image. The four later images present the same test case after applying the Resize function from PyTorch to pixelate the image and then reapplying the Resize function to restore it to its original size via bilinear interpolation. The MNIST, KMNIST, and Fashion MNIST images are represented in the data as negatives in the data (white writing on black background), and the reverse images are shown here. Additional information, including sources for datasets, appears in the notes to Table \ref{tab:datasets}.
\end{minipage}
\vspace{0.10 in}
\caption{First test image from each dataset at different resolution levels}
\label{fig:downsampled_images}
\end{figure*}

The ensemble approaches proposed in this study are tested on seven well-known image classification datasets from the computer vision literature. Five include thumbnail-sized images of digits from handwriting or house numbers, categories of Japanese characters, articles of clothing, and animals and vehicles. Model projections are generated using three different artificial neural networks as well as a logistic regression-based classifier. Additionally, twelve different established deep learning classifiers are applied to larger images from the 1,000-category ImageNet dataset and to a supplemental validation set to ImageNet that is meant to be less susceptible to overfitting.

The seven subfigures in Figure \ref{fig:downsampled_images} illustrate the original and downsampled representations of images from the first test image from each of the seven image recognition datasets used in this study. Descriptions of the datasets appear in Table \ref{tab:datasets}. The first image within each subfigure shows the original image. For each of the first three subfigures, the image is represented as a 28x28 grid of grayscale pixel intensities (numbers from 0 to 255), accounting for $28*28 = 784$ bytes of storage or bandwidth. For each of the next two subfigures, the full-sized image is represented as three 32x32 grids of red, blue, and green pixel intensities, accounting for $32*32*3 = 3,072$ bytes. The final two subfigures show sample images from ImageNet and ImageNet-V2. Images in the two datasets vary in size; the original sizes of the sample images shown are 500 x 375 and 375 x 500, both requiring $500*375*3 = 562,500$ bytes (roughly half a megabyte) of data. As per the standardized data transformations in PyTorch, every ImageNet image is resized so that the smaller side has 256 pixels, which reduces these image sizes to $562,500*(256^2/375^2) = 262,144.$

Within each subfigure, in the second column, each image's height and width are reduced by a factor of two via pixelation and then returned to original size via bilinear interpolation. Thus, only $784/4 = 196$ bytes are required to represent each of the images in the second column of the first three rows, and only $3,072/4 = 768$ bytes are required to represent each of the images in the second column of the last two rows. For the ImageNet and ImageNet-V2 images, the images in the second column require $262,144/4 = 65,536$ bytes to represent. The remaining columns show images wth increasing amounts of pixelation. The labels listed in the row headings describe the label for each image.

\begin{table}[ht!]
\renewcommand{\arraystretch}{2}
\centering
\resizebox{0.8\textwidth}{!}{
\begin{tabular}{p{0.12\textwidth}|p{0.17\textwidth}|p{0.14\textwidth}|p{0.07\textwidth}|p{0.10\textwidth}|p{0.10\textwidth}|p{0.08\textwidth}|p{0.20\textwidth}|p{0.15\textwidth}}
Dataset & Domain & Classes & Colors & Size & \shortstack{Training\\Cases} & \shortstack{Test/\\Validation\\Cases} & \shortstack{Models\\Estimated} & Source \\
\hline
Modified National Institute of Standards and Technology (MNIST) & Handwritten digits & $0,1,...,9$ & Grayscale & 28x28 & 60,000 & 10,000 & LeNet-5, Logistic with 100 factors & \cite{Lecun1998b} \\
\hline
Kujushiji MNIST (KMNIST) & Handwritten Japanese characters & o, ki, su, tsu, na, ha, ma, ya, re, wo & Grayscale & 28x28 & 60,000 & 10,000 & LeNet-5, Logistic with 100 factors & \cite{Clanuwat2018} \\
\hline
Fashion MNIST & Pictures of clothing items & t-shirt/top, trouser, pullover, dress, coat, sandal, shirt, sneaker, bag, ankle boot & Grayscale & 28x28 & 60,000 & 10,000 & LeNet-5, Logistic with 100 factors & \cite{Xiao2017} \\
\hline
Street View Housing Numbers (SVHN) & Pictures of cropped house numbers, sometimes with distracting adjacent information & $0,1,...,9$ & RGB & 32x32 & 73,257 & 26,032 & DLA-Simple, ResNet-18, LeNet-5, Logistic with 300 factors & \cite{Netzer2011} \\
\hline
Canadian Institute for Advanced Research (CIFAR-10) & Pictures of animals and vehicles in varying contexts and positions & plane, car, bird, cat, deer, dog, frog, horse, ship, truck & RGB & 32x32 & 50,000 & 10,000 & DLA-Simple, ResNet-18, LeNet-5, Logistic with 300 factors & \cite{Krizhevsky2014} \\
\hline
ImageNet Large Scale Visual Recognition Challenge 2012 (ILSVRC 2012, ImageNet) & Wide variety of classes of objects in different contexts and positions & 1,000 different classes of organisms (mostly animals), artifacts (\emph{e.g.}, instruments, structures, consumer goods), natural objects, and food & RGB & Edges ranging from 50 to 5,000 pixels with median image size of 375 x 500 & 1,281,167 & 50,000 & ResNeXt101-32x8d, Wide ResNet-101-2, EfficientNet-b0, EfficientNet-b7, ResNet-101, DenseNet-201, VGG-19-bn, MobileNet v3 small, MobileNet v3 large, GoogLeNet, Inception v3, AlexNet & \cite{Russakovsky2015} \\
\hline
ImageNet-V2 & New independent validation dataset generated for use with models developed using ImageNet training data.me as ImageNet & Same as ImageNet & RGB & Edges ranging from 50 to 5,000 pixels with median image size of 375 x 500 & No separate training cases; IVSLRC 2012 used for training & 10,000 & ResNeXt101-32x8d, Wide ResNet-101-2, EfficientNet-b0, EfficientNet-b7, ResNet-101, DenseNet-201, VGG-19-bn, MobileNet v3 small, MobileNet v3 large, GoogLeNet, Inception v3, AlexNet & \cite{Recht2019} \\
\end{tabular}}
\resizebox{0.8\textwidth}{!}{
\begin{tabular}{p{\linewidth}}
\vspace{-0.1 in}
\small Notes: This table summarizes the seven datasets used in this analysis. Sample images from these datasets can be seen in Figure \ref{fig:downsampled_images}. Of the three new validation datasets provided by \cite{Recht2019}, the ``matched frequency'' one is used so that the proportions in each category are the same as in the original data. For the purposes of this study, the sets of observations used for out-of-sample forecasting are referred to as the ``test sample'' for the MNIST, KMNIST, Fashion MNIST, SVHN, and CIFAR-10 datasets and as the ``validation sample'' for ImageNet and ImageNet-V2. The two types of samples are used identically. The reason for the distinction is that, while the holdout samples are typically referred to as test samples, the ImageNet dataset has associated with it a distinct unlabeled test sample that is not used in this analysis.
\end{tabular}}
\caption{Datasets used in Analysis}
    \label{tab:datasets}
\end{table}

As the examples in Figure \ref{fig:downsampled_images} illustrate, some of the full-sized representations in the first column are straightforward for a human to correctly classify with a high degree of confidence. The MNIST and SVHN images are easily discernable as the numbers 7 and 5, and the Fashion MNIST dataset clearly shows a heeled shoe or boot. The KMNIST dataset requires a classifier with knowledge of classical Japanese script, but the shape of the writing is easily identifable. The CIFAR-10 image is not immediately recognizable as a cat, but cat is the obvious choice among the options that also include plane, car, bird, deer, dog, frog, horse, ship, and truck. The images are still somewhat recognizable at the first two levels of downsampling, but with successively greater difficulty and less confidence, to the point that any classification of the 2x2 image would be a wild guess. The ImageNet and ImageNet-V2 images require some specialized knowledge to correctly classify even at full resolution. A causal observer could easily mistake the tench in Figure \ref{fig:imagenet} for another type of fish or believe that the sleeping bag in Figure \ref{fig:imagenet2} is a vest or a winter coat. Nevertheless, the distinguishing features of the images are mostly preserved after the first two levels of downsampling. At blurring beyond those two levels, however, the key features of the images are hard to discern by a human observer.

Table \ref{tab:datasets} describes the seven datasets from which the full-sized images in Figure \ref{fig:downsampled_images} are taken, including the categories to be learned, the features of the cases, and the models that are estimated on them. The first five cases involve small images that divide into 10 categories, and the last two involve larger images that divide into 1,000 categories.

\section{Classifier Specifications} \label{downsampling specifications}

The classification models used in this anaylsis are summarized in Table \ref{tab:nns}. The 16 models listed in the table include a logistic regression-based classifier that is used as a benchmark model as well as fifteen neural network approaches. The table lists the sources, numbers of parameters, and a brief description of key design elements for each of these classifiers. Additionally, the rightmost column lists the datasets for which the model is applied. Each model's parameters are fitted using the complete dataset of full-sized training images and then applied out-of-sample to classify the full-sized test or validation images and the pixelated variants. In each case, the downsampled images are resized in PyTorch as described in section \ref{downsampling data}. Three measures of model complexity are presented in the table. The number of trained parameters in the model captures the level of complexity of the modeled relationships between the features and labels observed in the training data. GMACs per case, a variation on Floating Point Operations (FLOPs), describes the amount of computation required to classify a single test case. GMACs was estimated using the fvcore package in PyTorch. Due to known limitations with profiling of this form, a hardware-based measure is presented as well---the number of miliseconds required to classify a test case, as measured from testing on a MacBook Pro.

The first few models are specific to the small image datasets. The logistic model takes as predictors the first 100 eigenvectors from the unscaled matrix of $28*28=784$ black-and-white pixel intensities for the MNIST-style images and the first 300 eigenvectors from the $32*32*3=3,072$ RGB pixel intensities for the CIFAR-10 and SVHN datasets.\footnote{Logistic regressions using more or fewer components yield similar results. The values of 100 and 300 were chosen because they had strong overall performance across the respective datasets and did not exhibit substantial amounts of overfitting.} In both cases, these principal components are used as predictors in 10 different logisitic regressions, one for each class. The coefficients are estimated via Maximum Likelihood Estimation (MLE). As the complexity measures show, the computation required per test case is negligible relative to the neural networks.

The second model, LeNet-5, is estimated on grayscale images, as in the original application to MNIST \cite{Lecun1998a}; the output layers are converted to propensities via softmax transforms; model parameters are trained over 15 epochs with a learning rate of 0.001.\footnote{Code-based implementation sourced from \cite{Lewinson2020}.}  The classifier resizes the 28x28 MNIST-style grayscale images to dimensions of 32x32 before they are fed into the network, and it converts the SVHN and CIFAR-10 images from 32x32 RGB to 32x32 grayscale.

The next two model implementations are designed for use on 32x32x3 RGB images and are consequently applied exclusively to the SVHN and CIFAR-10 datasets. For both the simplified Deep Layer Aggregation model and ResNet-18, the implementation is taken from \cite{Liu2022c}. multiple stages in the training process. Training for each proceeds in 200 epochs with a learning rate of 0.1. As with the LeNet-5, softmax is used to produce propensity scores.

\begin{table*}[ht!]
\renewcommand{\arraystretch}{1.25}
\centering
\resizebox{0.65\textwidth}{!}{
\begin{tabular}{p{0.18\textwidth}p{0.06\textwidth}p{0.06\textwidth}p{0.11\textwidth}p{0.10\textwidth}p{0.08\textwidth}p{0.45\textwidth}p{0.15\textwidth}}
\multirow{3}{*}{Model} & Design & Code & \multicolumn{3}{c}{Complexity} & \multirow{3}{*}{Key Design Elements} & \multirow{3}{*}{Applications} \\
\cline{4-6}
 & source & source & Param- & GMACs & Miliseconds & & \\
  & &  & eters (MM) & per case & per case & & \\
\hline
Logistic & \multicolumn{2}{c}{Standard} & 0.001 to 0.003 & 0.0000 & 0.0 & Top 100-300 principal components used as predictors & \multirow{2}{*}{\shortstack{MNIST, KMNIST, Fashion-\\MNIST, SVHN, CIFAR-10}} \\
LeNet-5 & \cite{Lecun1998a} & \cite{Lewinson2020} & 0.06 & 0.0004 & 0.1 & Early convolutional neural network & \\
\hline
DLA-Simple & \cite{Yu2018} & \multirow{2}{*}{\cite{Liu2022c}} & 15 & 0.92 & 5.8 & Aggregation of hidden layer outputs at multiple steps & \multirow{2}{*}{SVHN, CIFAR-10} \\
ResNet-18 & \cite{He2015} & & 11 & 0.56 & 4.8 & Feedback, depth & \\
\hline
ResNeXt-101-32x8d & \cite{Xie2017} & \multirow{12}{*}{\shortstack{Pre-\\trained\\para-\\meters\\from\\PyTorch\\\cite{Paszke2019}}} & 89 & 16.51 & 67.4 & Modular structure with repeated use of common transformation architecture & \multirow{12}{*}{\shortstack{ImageNet,\\ImageNet-V2}} \\
Wide ResNet 101-2 & \cite{Zagoruyko2017} & & 127 & 22.82 & 71.2 & Feedback, width & \\
EfficientNet-b0 & \cite{Tan2020} & & 5.3 & 0.40 & 25.5 & Coordinated scaling of width, & \\
EfficientNet-b7 &  & & 66 & 5.27 & 126 & depth, and resolution & \\
ResNet-101 & \cite{He2015} & & 45 & 7.85 & 29.6 & Feedback, depth & \\
DenseNet-201 & \cite{Huang2017} & & 20 & 4.37 & 35.6 & Dense connectivity across non-adjacent layers & \\
VGG-19-bn & \cite{Simonyan2015} & & 144 & 19.70 & 157 & Increased depth with small convolutional filters & \\
MobileNet v3 large & \cite{Howard2019} & & 5.5 & 0.23 & 15.0 & Network architecture search & \\
MobileNet v3 small & & & 2.5 & 0.06 & 10.7 & & \\
GoogLeNet & \cite{Szegedy2015a} & & 13 & 1.51 & 14.5 & Parallelism, increased depth and width & \\
Inception v3 & \cite{Szegedy2015b} & & 27 & 2.85 & 20.8 & Parallelism, expanded convolution and regularization & \\
AlexNet & \cite{Krizhevsky2014} & & 61 & 0.92 & 6.8 & Computational parallelism & \\
\hline
\end{tabular}}
\resizebox{0.65\textwidth}{!}{
\begin{tabular}{p{\linewidth}}
\vspace{-0.15 in}
\small Notes: This table describes the 16 classifiers used in this study, including the source for the classifier design and code-based implementation, the number of parameters estimated, key features of the design, and the datasets to which that classifier is applied in the current study. For each model, the parameters are fitted using the complete training dataset of full-sized images. The same parameters are applied to the original images from the test or validation dataset as well as to downsampled variants of those images in the image resolution application. GMACs denotes Giga-Multiply-Accumulate operations per case, a measure of computational complexity provided in the fvcore package in Python and a variation on Floating Point Operations (FLOPs). The GMAC values shown here denote the cost of projection and do not include the cost of training. Miliseconds per test case are estimated by timing the projection of 10,000 test cases on a MacBook Pro (Apple M1 Pro chip).
\end{tabular}}
\caption{Summary of classifiers used in this study}
    \label{tab:nns}
\end{table*}

For each of the remaining 12 models, the pre-trained weights were obtained from PyTorch \cite{Paszke2019} and then applied to the validation samples from ImageNet and ImageNet-V2. The classification problem in the ImageNet and ImageNet-V2 datasets is more difficult than with the smaller images. The model must process larger and varying amounts of input data and select among 1,000 potential labels. As the ``Parameters'' column shows, the neural networks that are used to classify ImageNet-style images are generally more complex than those used to classify the smaller images, with numbers of parameters in some cases exceeding 100 MM---as compared to 60,000 for LeNet-5. Following the PyTorch documentation, a standard set of transformations is applied to the images: each is first resized to 256 (\emph{i.e.}, the smaller of the two edges is resized to 256 pixels, and larger side is resized so that the aspect ratio is unchanged), then center-cropped to 224, then normalized with RGB channel-specific means of 0.45, 0.456, and 0.406 and standard deviations of 0.229, 0.224, and 0.225. To test the models on downsampled variants of the images, an additional step is included in the transformation to resize to 128, 64, 32, or 16 prior to resizing to 256.

\section{Descriptive Results} \label{downsampling descriptive}

\subsection{Downsampling and Accuracy} \label{downsampling and accuracy}

Before presenting results from the sequential threshold-based model, Table \ref{tab:performance} shows how the classifiers used in the component steps perform on their own. No threshold or sequential approach is used; the logistic approach and the 15 neural networks are applied in each case to the complete sets of full-resolution training images, full-sized test or validation images, and downsampled test or validation images, where the downsampling rate varies across the columns. The values in the table illustrate the percentage of cases in which the correct class was chosen \. For each of the MNIST-style and small RGB image datasets, the number of classes is 10, so that pure random chance would produce a rate of accuracy of 10\%. For the ImageNet and ImageNet-V2 datasets, the number of classes is 1,000, and the Top-1 accuracy is shown, so that pure random chance would produce a rate of accuracy of 0.1\%. Results for the classification of MNIST-style images appear in panel A, panel B shows results for the classification of small RGB images, and panel C shows results for the larger RGB images from ImageNet and ImageNet-V2. For the benchmark cases in which the full-sized images are used, the neural networks used here are not those that achieve peak state-of-the-art performance but are representative of the sorts of networks commonly seen in the literature. Additionally, some weaker classifiers---the logistic regression approach as well as the LeNet-5 applied to grayscale versions of the RGB images---are included for comparison in order to illustrate how the proposed strategy performs across a variety of situations.\footnote{Recent studies report performance of 99.91\% on MNIST, 99.15\% on KMNIST, 96.91\% on Fashion MNIST, 99.01\% on SVHN, and 99.5\% on CIFAR-10 \cite{An2020,DipuKabir2022,Tanveer2020,Foret2021,Dosovitskiy2021}. Thus, the LeNet-5's performance on the full-sized test images, and the DLA-Simple and ResNet-18 performance on the SVHN datasets, as shown in Table \ref{tab:performance}, are somewhat below the state-of-the-art rates of accuracy, although the performance of the DLA-Simple and ResNet-18 classifiers on the CIFAR-10 dataset is comparable to what has been achieved in the literature. On the ImageNet validation dataset, recent very large models (with $\sim$ 2 billion parameters) have been able to achieve Top 1 accuracy rates at or just below 91\% by training a multi-modal model that supplements image data with text-based information from captions \cite{Yu2022} or by averaging forecasts across multiple large models \cite{Wortsman2022}.} The different rows illustrate how the performance varies across datasets and models.

\begin{table}
\renewcommand{\arraystretch}{1.1}
\centering
\resizebox{0.65\textwidth}{!}{
\begin{tabular}{cccccccc}
\multicolumn{8}{c}{Panel A: MNIST-style images} \\
\multirow{2}{*}{Dataset} & \multirow{2}{*}{Model} & Training & \multicolumn{5}{c}{Test} \\
 & & 28x28 & 28x28 & 14x14 & 7x7 & 4x4 & 2x2 \\
\hline
MNIST  & Logistic & 91.2\% & 91.5\% & 91.0\% & 78.4\% & 47.8\% & 14.9\% \\
 & LeNet-5 & 99.5\% & 98.5\% & 97.1\% & 60.1\% & 17.0\% & 9.5\% \\

KMNIST  & Logistic & 81.5\% & 67.9\% & 67.5\% & 61.4\% & 43.9\% & 19.2\% \\
 & LeNet-5 & 99.1\% & 92.7\% & 90.7\% & 69.8\% & 30.4\% & 9.6\% \\
Fashion MNIST  & Logistic & 85.1\% & 83.8\% & 80.2\% & 72.7\% & 46.3\% & 24.1\% \\
 & LeNet-5 & 91.5\% & 87.7\% & 85.3\% & 72.3\% & 35.8\% & 17.8\% \\
 \\
\multicolumn{8}{c}{Panel B: Small RGB images} \\
\multirow{2}{*}{Dataset} & \multirow{2}{*}{Model} & Training & \multicolumn{5}{c}{Test} \\
 & & 32x32 & 32x32 & 16x16 & 8x8 & 4x4 & 2x2 \\
\hline
SVHN & Logistic & 28.5\% & 24.7\% & 24.4\% & 23.1\% & 19.2\% & 17.5\% \\
& LeNet-5 (Grayscale) & 91.3\% & 83.1\% & 82.2\% & 72.3\% & 28.3\% & 18.5\% \\
 & DLA-Simple & 100.0\% & 97.3\% & 96.2\% & 84.8\% & 28.6\% & 17.7\% \\
 & ResNet-18 & 100.0\% & 97.2\% & 96.2\% & 83.7\% & 27.2\% & 18.1\% \\
CIFAR-10 & Logistic & 42.1\% & 41.0\% & 41.0\% & 40.2\% & 35.6\% & 26.2\% \\
 & LeNet-5 (Grayscale) & 56.6\% & 49.7\% & 46.4\% & 32.1\% & 20.4\% & 15.8\% \\
 & DLA-Simple & 100.0\% & 94.9\% & 26.0\% & 13.4\% & 11.9\% & 11.3\% \\
 & ResNet-18 & 100.0\% & 95.4\% & 27.7\% & 20.5\% & 18.2\% & 14.8\% \\
 \\
\multicolumn{8}{c}{Panel C: Larger RGB images (size = smaller edge)} \\
\multirow{2}{*}{Dataset} & \multirow{2}{*}{Model} & Training & \multicolumn{5}{c}{Validation} \\
 & & 256 & 256 & 128 & 64 & 32 & 16 \\
\hline
ImageNet & ResNeXt-101-32x8d & 95.7\% & 79.3\% & 72.2\% & 52.2\% & 22.9\% & 5.1\% \\
& Wide ResNet 101-2 & 96.5\% & 78.8\% & 70.7\% & 49.4\% & 20.2\% & 4.7\% \\
& EfficientNet-b0 & 91.2\% & 77.7\% & 69.4\% & 48.1\% & 21.2\% & 4.6\% \\
& EfficientNet-b7 & 94.0\% & 73.9\% & 64.6\% & 48.8\% & 24.7\% & 6.6\% \\
& ResNet-101 & 90.4\% & 77.4\% & 70.6\% & 51.0\% & 21.9\% & 4.4\% \\
& DenseNet-201 & 89.5\% & 76.9\% & 69.1\% & 50.9\% & 21.9\% & 4.2\% \\
& VGG-19-bn & 84.8\% & 74.2\% & 64.1\% & 37.4\% & 12.6\% & 2.5\% \\
& MobileNet v3 large & 93.1\% & 67.7\% & 54.9\% & 37.4\% & 14.4\% & 2.8\% \\
& MobileNet v3 small & 79.9\% & 74.1\% & 65.8\% & 44.0\% & 16.4\% & 3.1\% \\
& GoogLeNet & 77.5\% & 69.8\% & 60.7\% & 38.9\% & 11.6\% & 1.6\% \\
& Inception v3 & 81.2\% & 69.5\% & 61.3\% & 42.6\% & 15.3\% & 2.6\% \\
& AlexNet & 73.3\% & 56.6\% & 44.8\% & 25.9\% & 8.6\% & 2.3\% \\
ImageNet V2 & ResNeXt-101-32x8d & & 67.5\% & 59.7\% & 41.7\% & 18.0\% & 4.1\% \\
(Matched & Wide ResNet 101-2 & & 66.5\% & 58.1\% & 39.0\% & 15.9\% & 3.6\% \\
Frequency) & EfficientNet-b0 & & 65.7\% & 56.2\% & 37.5\% & 16.5\% & 3.4\% \\
& EfficientNet-b8 & & 61.8\% & 52.5\% & 39.0\% & 19.3\% & 5.2\% \\
& ResNet-101 & & 65.6\% & 58.0\% & 40.5\% & 16.8\% & 3.5\% \\
& DenseNet-201 & & 64.7\% & 57.0\% & 40.6\% & 17.3\% & 2.8\% \\
& VGG-19-bn & & 61.9\% & 51.4\% & 29.2\% & 9.5\% & 1.9\% \\
& MobileNet v3 large & & 54.7\% & 43.0\% & 29.1\% & 10.9\% & 2.2\% \\
& MobileNet v3 small & & 60.5\% & 52.4\% & 34.2\% & 12.4\% & 2.4\% \\
& GoogLeNet & & 57.9\% & 48.7\% & 30.9\% & 8.9\% & 1.2\% \\
& Inception v3 & & 57.6\% & 49.5\% & 33.4\% & 12.0\% & 1.9\% \\
& AlexNet & & 43.5\% & 33.7\% & 19.9\% & 7.0\% & 1.7\% \\
\hline
\end{tabular}}
\resizebox{0.75\textwidth}{!}{
\begin{tabular}{p{\linewidth}}
\vspace{-0.1 in}
\small Notes: Each row presents the percent of cases correctly classified for a different dataset-model combination. In each case, the input data is a matrix of pixel intensities for an image, and the task is to assign the case to one of ten possible categories. Training is performed separately for each dataset-model combination, and the same trained model is then applied to full-resolution and pixelated test images. Additional details about the datasets and classifiers are presented in Table \ref{tab:datasets} and Section \ref{downsampling specifications}.
\end{tabular}}
\caption{Accuracy by Dataset, Model, Sample, and Image Resolution} 
\label{tab:performance}
\end{table}

As the results from the first panel show, when the full training or test images are used, the LeNet-5 accurately classifies MNIST-style images 87.7\% to 99.5\% of the time. The accuracy is somewhat lower for the simpler logistic regression approach, which correctly classifies cases 67.9\% to 91.5\% of the time. Classifications are most accurate for the digit identification task of MNIST and less for the more complex problems of identifying classes of Japanese characters in KMNIST or types of clothing in Fashion MNIST. Performance declines as we move from left to right, and increasing amounts of downsampling are applied to the images, and at the very right, when the 2x2 images are used, only 9.5\% to 24.1\% of cases are correctly classified.

For the small RGB images in panel B, the classifiers exhibit more varied performance on the full-sized training and test images. The LeNet-5 performs well on the grayscale SVHN images, correctly classifying 91.3\% of training cases and 83.1\% of test cases. The logistic regression model performs far worse, with accuracy of 28.5\% and 24.7\% on the same two samples. One potential explanation for the worse performance of the logistic classifier is that, due to the varying colors and distracting adjacent information in the images, the representations are more complex than can be captured effectively with the first 300 principal components. For the CIFAR-10 images, both the LeNet-5 and logistic classifiers perform poorly, correctly recognizing the objects in only 41.0\% to 56.6\% of the full-sized training and test cases---possibly due to the complexity of the representations and the importance of color for identification. The two deeper and more sophisticated neural networks---the DLA-Simple and ResNet-18 models---perform far better on both the SVHN and CIFAR-10 images, correctly classifying 100.0\% of the full-sized training images and 94.9\% to 97.3\% of the full-sized test images. As in panel A, the classifiers' performance in panel B declines substantially as we move rightward to increasing amounts of downsampling. The drop is most severe for the CIFAR-10 images. The DLA-Simple and ResNet-18 perofrmance drops from 94.9\% and 95.4\% on the full-sized test images to 26.0\% and 27.7\% on the 16x16 images. This sharp decline is consistent with the intuitive observation from Figure \ref{fig:cifar} that the CIFAR-10 images are difficult to classify even with at full resolution. Due to the varying positions and contexts in which the same category of animal or vehicle can be presented, accurate classification appears to require both the complexity of a deep neural network and the granular data of a full-resolution image.

The models listed in panel C are shown in roughly decreasing order of validation accuracy, but with models from the same family (EfficientNet and MobileNet) presented together. The models are the same as those shown in \cite{Rohlfs2022d}, and as that study and \cite{Recht2019} both note, a considerable amount of overfitting can be seen in these models, evidenced both by the degradation of performance moving from training to validation and the further degradation moving from ImageNet to ImageNet-V2. Thus, the top-performing model in the group, ResNeXt-101-32x8d, achieves 95.7\% accuracy among the training cases but only 79.3\% accuracy among validation cases in ImageNet and 67.5\% accuracy among validation cases in ImageNet-V2. Among the set of 12, the lowest performer is the relatively early AlexNet model, which achieves 73.3\% training accuracy, 56.6\% validation accuracy on ImageNet, and 43.5\% validation accuracy on ImageNet-V2. For all 12 of the models, both when applied to ImageNet and to ImageNet-V2, downsampling the images causes substantial degradation in performance. Downsampling to 128 and 64 still produces reasonable rates of accuracy, however. And even at the extreme amount of downsampling down to 16, the classifiers achieve rates of accuracy from 1.2\% to 5.2\%, considerably higher than the 0.1\% that would be obtained due to random chance.

\subsection{Propensity Scores} \label{downsampling propensity}

\begin{figure*}
\centering
\begin{subfigure}{0.4\textwidth}
  \centering
  \includegraphics[width=\textwidth]{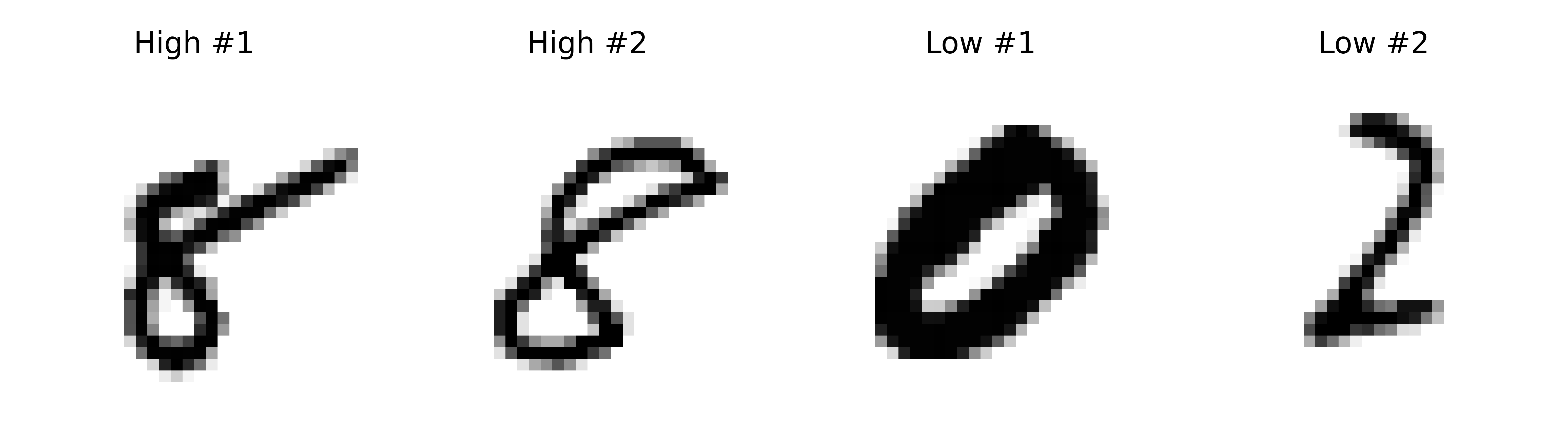}
  \caption{MNIST, Labels = 8, 8, 0, 2}
  \label{fig:mnist_bestworst}
\end{subfigure}
\begin{subfigure}{0.4\textwidth}
  \centering
  \includegraphics[width=\textwidth]{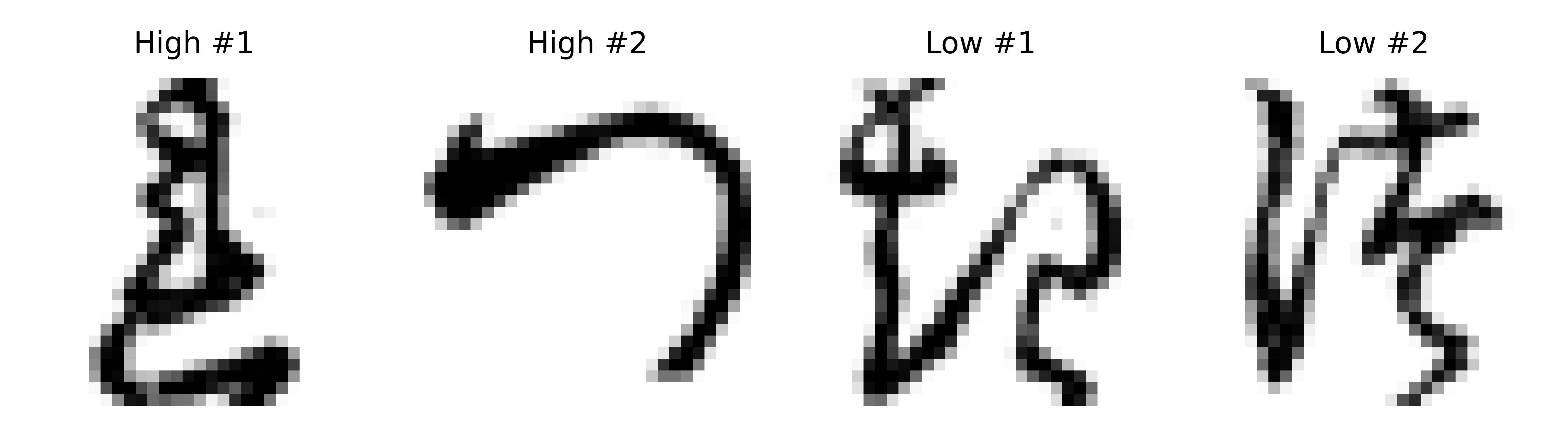}
  \caption{KMNIST, Labels = ``re,'' ``tsu,'' ``ki,'' ``tsu''}
  \label{fig:kmnist_bestworst}
\end{subfigure}
\begin{subfigure}{0.4\textwidth}
  \centering
  \includegraphics[width=\textwidth]{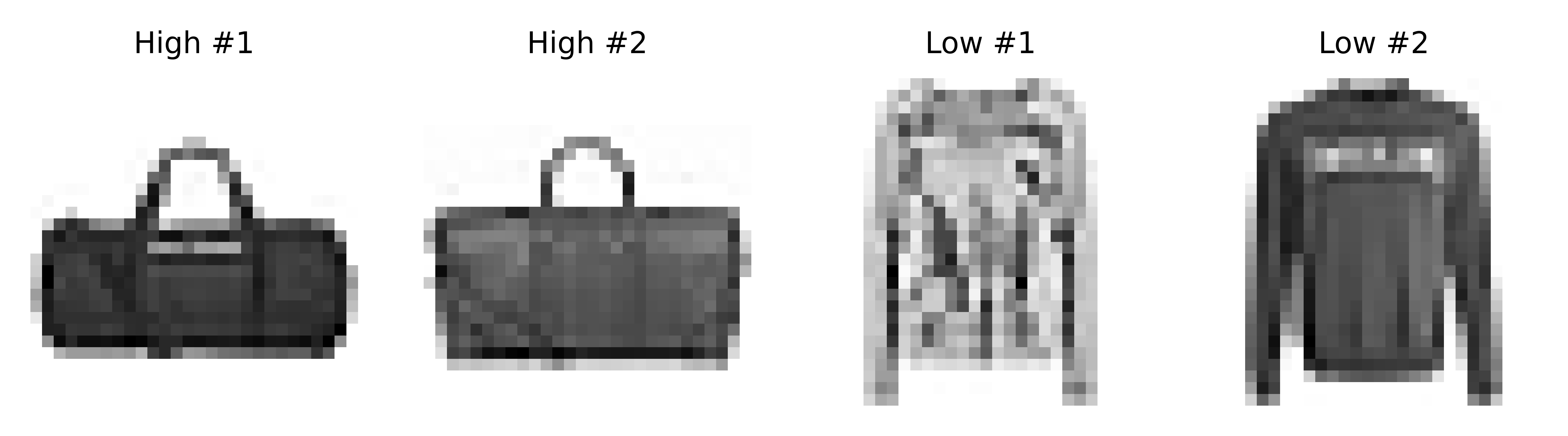}
  \caption{Fashion MNIST, Labels = ``bag,'' ``bag,'' ``pullover,'' ``pullover''}
  \label{fig:fashionmnist_bestworst}
\end{subfigure}
\begin{subfigure}{0.4\textwidth}
  \centering
  \includegraphics[width=\textwidth]{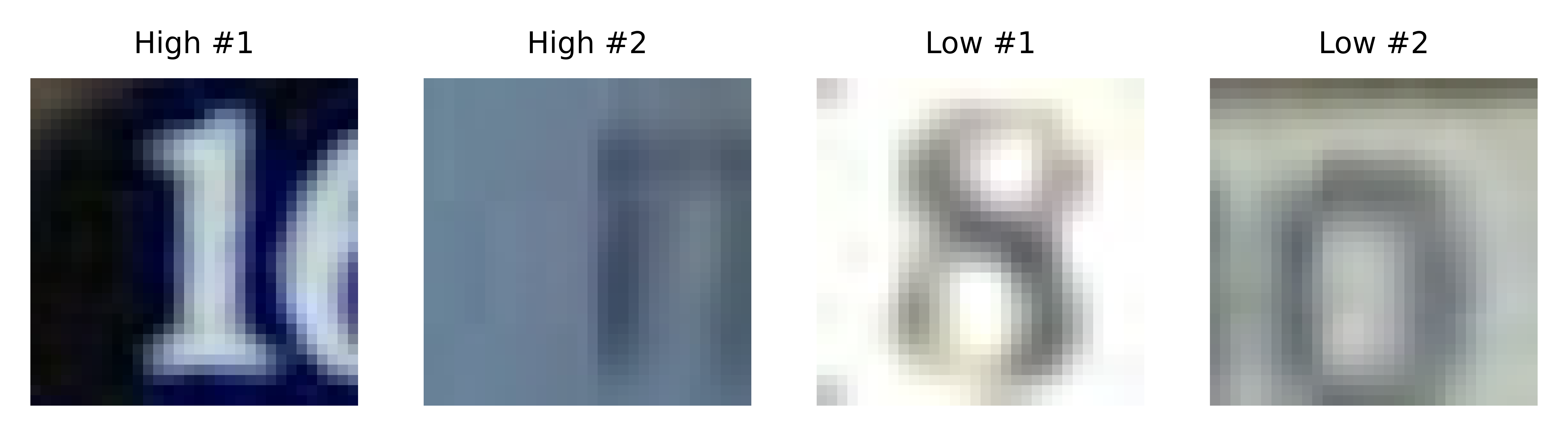}
  \caption{SVHN, Labels = 1, 1, 8, 0}
  \label{fig:svhn_bestworst}
\end{subfigure}
\begin{subfigure}{0.4\textwidth}
  \centering
  \includegraphics[width=\textwidth]{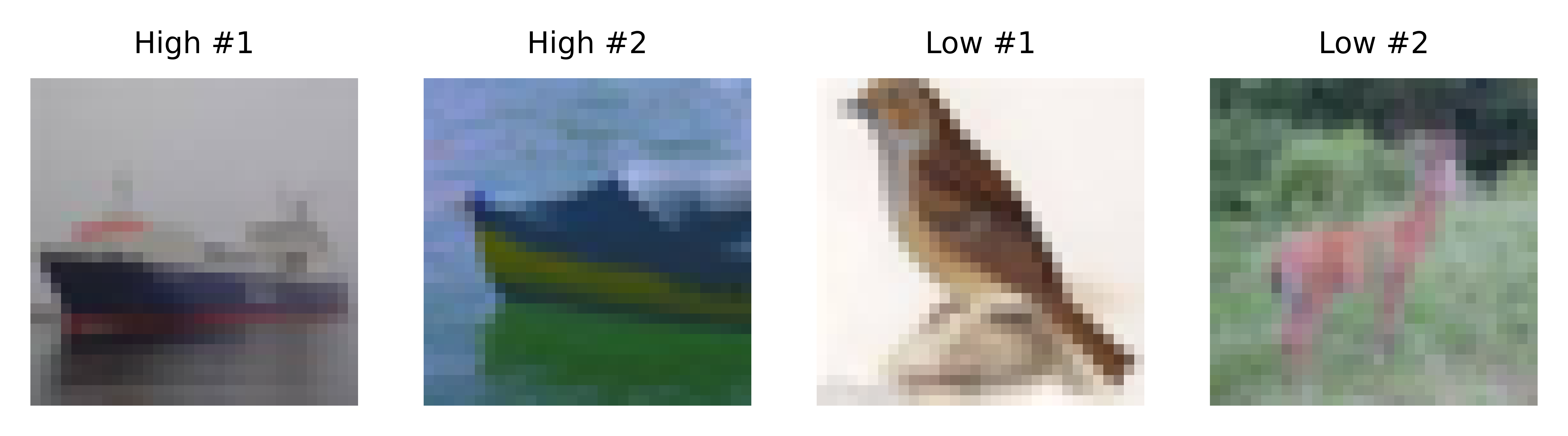}
  \caption{CIFAR-10, Labels = ``ship,'' ``ship,'' ``bird,'' ``deer''}
  \label{fig:cifar_bestworst}
\end{subfigure}
\begin{subfigure}{0.4\textwidth}
  \centering
  \includegraphics[width=\textwidth]{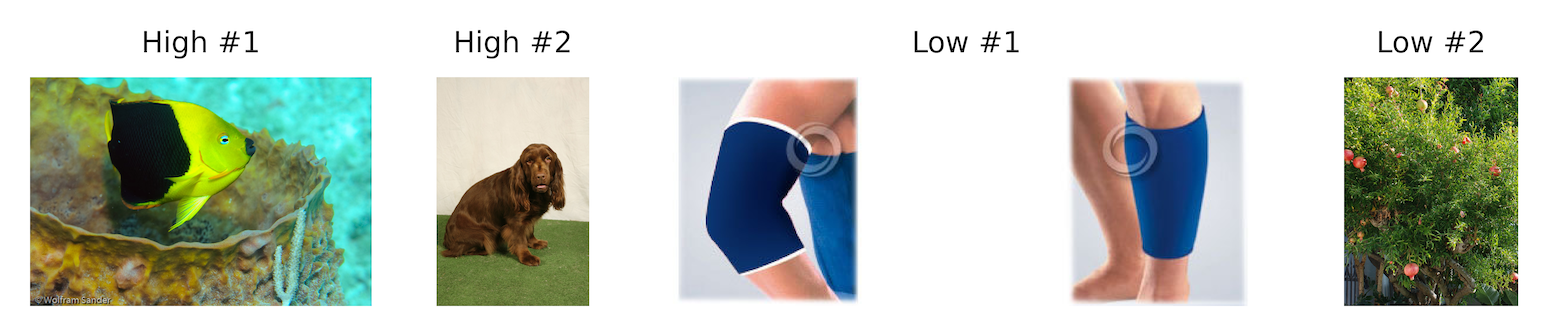}
  \caption{ImageNet, Labels = ``rock beauty,'' ``cocker spaniel,'' ``knee pad,'' ``pomegranate''}
  \label{fig:imagenet_bestworst}
\end{subfigure}
\begin{subfigure}{0.4\textwidth}
  \centering
  \includegraphics[width=\textwidth]{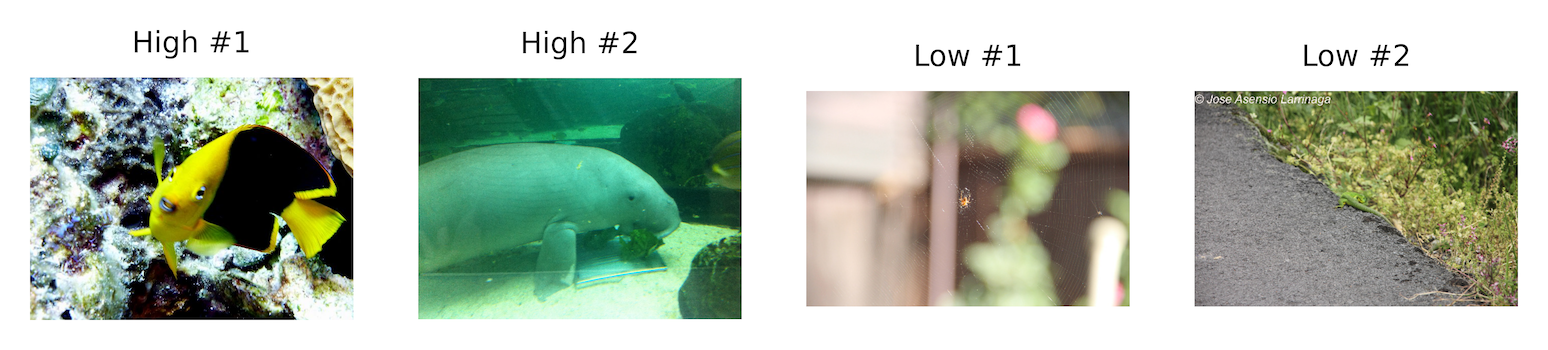}
  \caption{ImageNet-V2, Labels = ``rock beauty,'' ``dugong,'' ``spider web,'' ``green lizard''}
  \label{fig:imagenetv2_bestworst}
\end{subfigure}
\begin{minipage}{0.9\linewidth}
\vspace{0.10 in}
\flushleft \small Notes: Within each of the seven panels, the first two images show ``easy'' cases that can be correctly classified based upon downsampled data, and the second two images show ``difficult'' cases for which a network-based classifier has a low level of confidence when using downsampled data. The two easiest cases for each dataset are chosen from among those for which the classifier produced a correct label based upon an image that was downsampled by a factor of four in each direction (as in the middle images in each subfigure of Figure \ref{fig:downsampled_images}). From that set, the cases with the highest propensity scores were selected. The ``difficult'' cases for each dataset are those with the lowest propensity scores, regardless of whether they were correctly or incorrectly classified. For the MNIST-style images, the propensity scores and classifications are taken from the projections from the LeNet-5 model. For SVHN and CIFAR-10, they are averaged between the DLA-Simple and ResNet-18 projections, where the ``easy'' images are chosen from cases that were correctly classified by both models. For the ImageNet and ImageNet-V2 images, the propensity scores are averaged across the 12 classifiers used on those data in the current paper. In both of those datasets, the cases with the top 2 propensity scores are correctly classified by all 12 models. Additional information on the data sources and classification models appears in Tables \ref{tab:datasets} and \ref{tab:nns}.
\end{minipage}
\vspace{0.10 in}
\caption{Easy and Difficult Examples from Each Dataset}
\label{fig:bestworst}
\end{figure*}

To better understand the manner in which the threshold-based approach discriminates images, Figure \ref{fig:bestworst} presents example images that are ``easy''---and particularly likely to be classified correctly in the first stage, when the downsampled variants are used as inputs---and examples that are ``difficult'' and require more thorough analysis. The two images on the left of each subfigure are cases for which the neural network correctly labels that image from the downsampled data and assigns a high propensity score to its classification. The two images on the right of each subfigure are cases for which the classifier assigns a low propensity score from the downsampled image. Examining these cases helps to illustrate what characteristics of the images are most likely to confound the classifier and the extent to which problem difficulty varies across cases in each dataset. As can be seen from the examples shown here, some categories and styles of images rely more heavily upon the granular features of the highest possible resolution, while others can easily be discerned from blurred versions. The distinction is especially pronounced for the ImageNet and ImageNet-V2 images and to some extent in CIFAR-10; the more difficult cases from those datasets are difficult to discern for even a human classifier, while the easier cases are unmistakable. As some of the examples from the smaller-sized images illustrate, the level of difficulty is associated in some cases with the image label, and when the classifiers are applied to blurred images, they tend to favor some labels over others.

The eight digits---the four MNIST images in Figure \ref{fig:mnist_bestworst} and the four SVHN images in Figure \ref{fig:svhn_bestworst}---are generally straightforward for a human to classify. The two ``easy'' cases for MNIST are both the number eight, and the two easy cases for SVHN are both the number one. The second easy case for the SVHN is faint but has a clean centrally placed vertical line, and while there are distracting elements on the right-hand side, they do not have a form that would increase the likelihood of a competing label like seven or four. Among the two MNIST images that are classified with low confidence, the zero is misclassified as an eight, and the two is misclassified as a four. Among the two SVHN images with low confidence classifications, the eight is misclassified as a one by both DLA-Simple and ResNet-18, and the zero is misclassified as a three by DLA-Simple and as a four by ResNet-18. In general, the models exhibit bias when applied to the downsampled images, with LeNet-5 overclassifying downsampled MNIST images as eights, DLA-Simple overlassifying downsampled SVHN images as ones, threes, and fours, and ResNet-18 overclassifying downsampled SVHN images as ones and fours.

Among the KMNIST images in Figure \ref{fig:kmnist_bestworst}, the two difficult and misclassified images are noticeably more intricate than the easy images. The second difficult image, whose correct class is ``tsu,'' is misclassified by LeNet-5 as ``ki,'' which is the class of the first difficult image. The two easy Fashion MNIST images in Figure \ref{fig:fashionmnist_bestworst} are in the ``bag'' category, and the two difficult cases are in the ``pullover,'' category. The first of the two pullovers is incorrectly classified as a ``t-shirt/top,'' and the second is correctly classified as a pullover but with a low propensity score of 24.5\%. As with the MNIST and SVHN cases, there is some bias in the LeNet-5 classifications of the downsampled KMNIST and Fashion MNIST images. For the downsampled Fashion MNIST images, the classifier favors ``t-shirt/top'' and ``bag.''

The two easy CIFAR-10 images in Figure \ref{fig:cifar_bestworst} fall into the ``ship'' category, which is one of the labels favored by the ResNet-18 classifier when applied to the downsampled data. In addition to the straight lines, the blue color in the second case probably helps in this identification. The first of the two difficult cases is correctly classified as a bird by DLA-Simple but is misclassified as a cat by ResNet-18. The second of the difficult cases, whose true label is ``deer,'' is misclassified as a ``dog'' by both DLA-Simple and ResNet-18.

In the ImageNet and ImageNet-V2 cases in Figures \ref{fig:imagenet_bestworst} and \ref{fig:imagenetv2_bestworst}, the highest scoring ``easy'' image in both datasets is a ``rock beauty'' anglefish, which has a distinctive yellow and black coloration. Among the other two easy images are a cocker spaniel in an easily identifiable pose and a ``dugong,'' a relative of the manatee whose smooth features are unlikely to change markedly with image downsampling. The most difficult ImageNet validation case is a knee pad that appears in the data as two pictures separated by a large white space---an image that might be difficult for a typical human to classify. That case was misclassified by 5 of the 12 models as a ``corkscrew,'' and the other 7 models misclassified it as ``can opener,'' ``letter opener,'' ``envelope,'' ``face powder,'' ``website,'' ``hog,'' and ``nematode'' (a type of roundworm). The other three difficult cases---the second lowest propensity case from ImageNet and the two lowest propensity cases from ImageNet-V2---are those for which a high level of resolution is essential for understanding what is shown. The pomegranates in the second-lowest propensity ImageNet image are hard to identify among the leaves and would be lost in the blurring of a lower resolution image. Of the 12 classifiers, six mislabeled the image as a ``croquet ball,'' three mislabeled it as a ``worm fence'' (a zig-zagging style of fence), and the other three mislabeled it as a ``cardoon'' (a type of thistle), a ``black stork,'' and a ``greenhouse.'' Among the difficult ImageNet-V2 cases, the hard-to-spot spider web in the first case is mistaken by 4 of the 12 classifiers to be a ``chime'' and is also misclassified as ``jacamar'' (a type of bird), ``patas'' (a type of monkey), ``lipstick,'' ``rose hip'' (a type of fruit), ``picket fence,'' ``worm fence,'' ``prison,'' and ``tobacco shop.'' While all 12 models correctly surmise that the last image is that of an animal, the well-camouflaged green lizard is mistaken by two for a ``curly-coated retriever,'' two misclassify it as a ``skunk,'' two as a ``gray fox,'' and the others mislabel it as ``axolotl'' (a type of salamander), ``guenon'' (a type of monkey), ``toy terrier,'' ``toy poodle,'' ``standard schnauzer,'' and ``fox squirrel.''

\begin{figure*}[t]
  \centering
\begin{subfigure}{\textwidth}
  \centering
  \includegraphics[width=\textwidth]{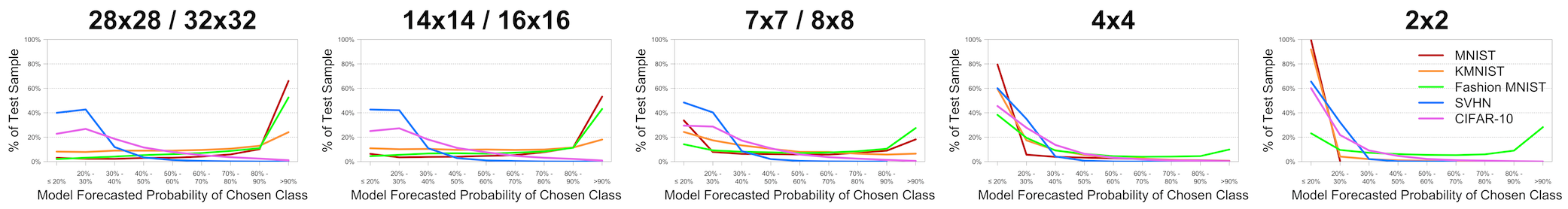}
  \caption{Logistic}
  \label{fig:count_logistic}
\end{subfigure}
\begin{subfigure}{\textwidth}
  \centering
  \includegraphics[width=\textwidth]{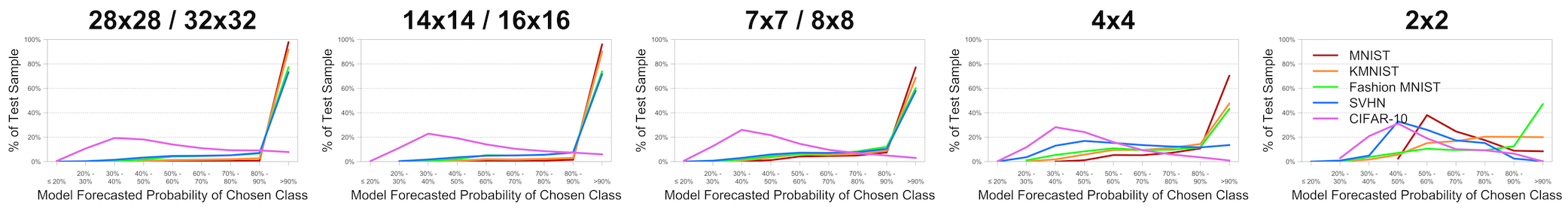}
  \caption{LeNet-5}
  \label{fig:count_lenet}
\end{subfigure}
\begin{subfigure}{\textwidth}
  \centering
  \includegraphics[width=\textwidth]{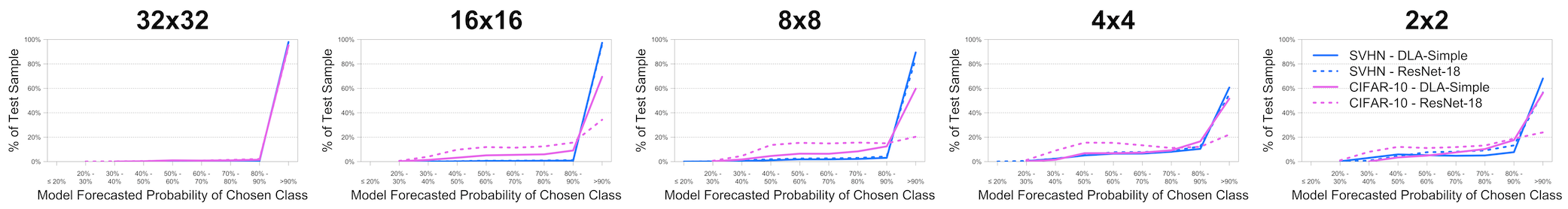}
  \caption{DLA-Simple and ResNet-18}
  \label{fig:count_dlares}
\end{subfigure}
\begin{subfigure}{\textwidth}
  \centering
  \includegraphics[width=\textwidth]{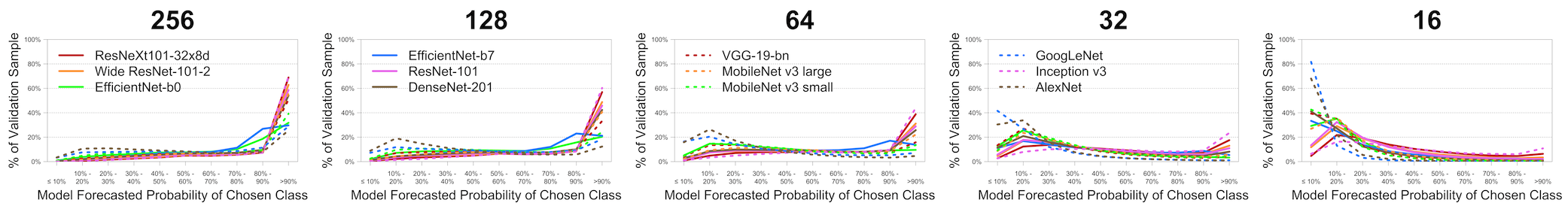}
  \caption{ImageNet}
  \label{fig:count_imagenet}
\end{subfigure}
\begin{subfigure}{\textwidth}
  \centering
  \includegraphics[width=\textwidth]{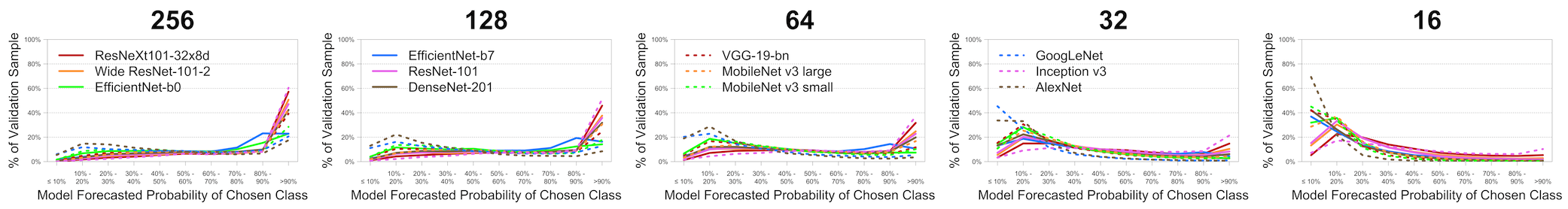}
  \caption{ImageNet-V2}
  \label{fig:count_imagenetv2}
\end{subfigure}
\begin{minipage}{\linewidth}
\vspace{0.10 in}
\flushleft \small Notes: See notes to Table \ref{tab:performance}. Each of these 25 plots shows histograms of the percentage of cases in the test or validation sample for which the modeled level of confidence in the chosen classification (the internal propensity score) falls within the interval shown on the horizontal axis. Large positive values on the right-hand side of the graph indicate large proportions of observations for which the model is highly confident in its classification, and large positive values on the left-hand side indicate large proportions of observations for which the model has low confidence in its classification. The first three rows of figures correspond to different models, the columns correspond to different rates of downsampling, and the different lines within each chart correspond to the different datasets. The latter two rows show results for all 12 of the models applied to the ImageNet and ImageNet-V2 datasets, separately, with the different lines within each chart corresponding to the different classifiers.
\end{minipage}
\vspace{-0.02 in}
\caption{Histograms of Modeled Propensity Scores}
\label{fig:histogram}
\end{figure*}

\begin{figure*}
\begin{subfigure}{\textwidth}
  \centering
  \includegraphics[width=\textwidth]{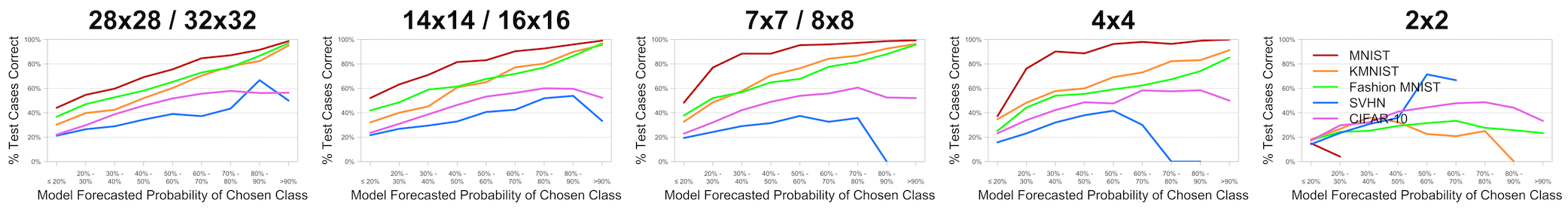}
  \caption{Logistic}
  \label{fig:correct_logistic}
\end{subfigure}
\begin{subfigure}{\textwidth}
  \centering
  \includegraphics[width=\textwidth]{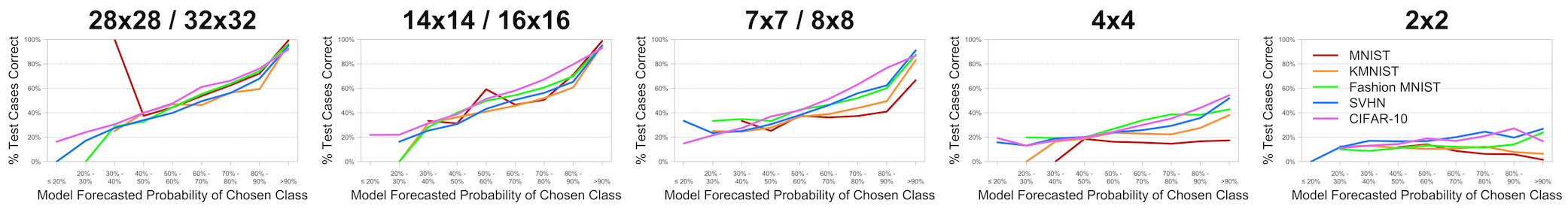}
  \caption{LeNet-5}
  \label{fig:correct_lenet}
\end{subfigure}
\begin{subfigure}{\textwidth}
  \centering
  \includegraphics[width=\textwidth]{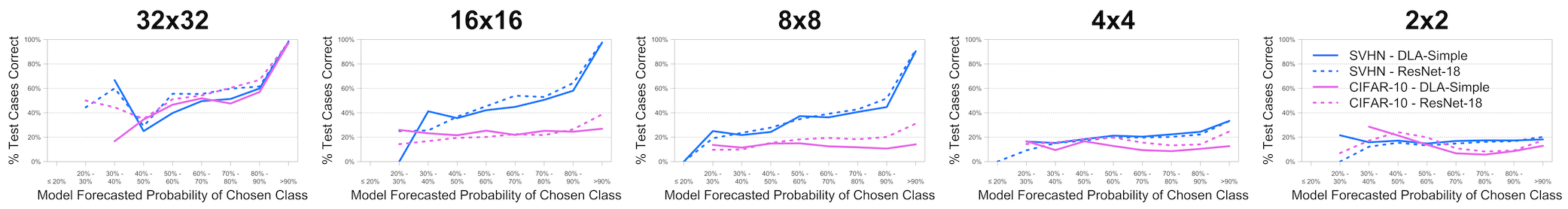}
  \caption{DLA-Simple and ResNet-18}
  \label{fig:correct_dlares}
\end{subfigure}
\begin{subfigure}{\textwidth}
  \centering
  \includegraphics[width=\textwidth]{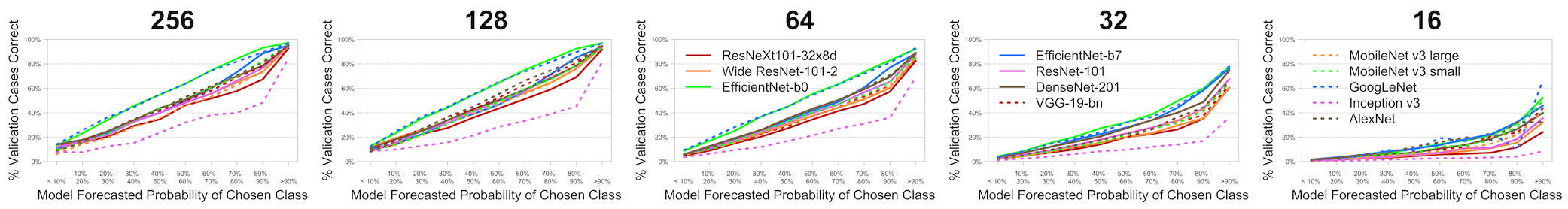}
  \caption{ImageNet}
  \label{fig:correct_imagenet}
\end{subfigure}
\begin{subfigure}{\textwidth}
  \centering
  \includegraphics[width=\textwidth]{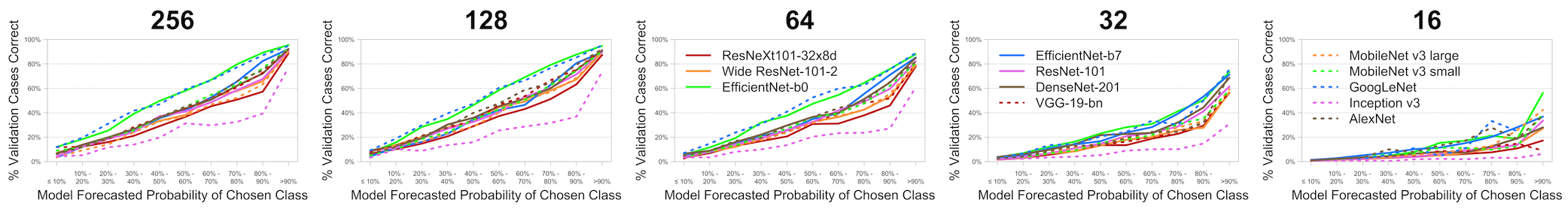}
  \caption{ImageNet-V2}
  \label{fig:correct_imagenetv2}
\end{subfigure}
\begin{minipage}{\linewidth}
\vspace{0.10 in}
\flushleft \small Notes: See notes to Figure \ref{fig:histogram} and Table \ref{tab:performance}. These plots follow the same structure as those in Figure \ref{fig:histogram}, but the variable plotted along the horizontal axis is classification accuracy rather than percentage of the sample. The charts illustrate the extent to which the models' internal propensity scores are predictive of test performance. For a case in which the score is a perfect proxy for classification accuracy, the plot would show a straight upward-sloping line from roughly 15\% on the left-hand side to 95\% on the right-hand side. In order for the propensity scores to be useful for the sequential classifier, however, the magnitude of the slope and the scale of the propensity values are less important than the existence of a reliably positive sign indicating that the score is informative about model performance on a case-by-case basis.
\end{minipage}
\vspace{-0.02 in}
\caption{Accuracy by Modeled Propensity Scores}
\label{fig:correct}
\end{figure*}

Next, Figures \ref{fig:histogram} and \ref{fig:correct} further explore the models' internal propensity scores by illustrating their distributions and accuracy in the test data. In each of the 25 charts in Figure \ref{fig:histogram}, the test cases from each dataset are divided into nine categories based upon the models internal propensity scores for their chosen classifications for those cases. The categories are plotted along the horizontal axis in ranges of ten percentage points. For the MNIST-style and small RGB problems that have ten possible labels, as shown in Figures \ref{fig:count_logistic} through \ref{fig:count_dlares}, these groups of propensity scores range from $\le 20\%$, $20\%-30\%$, $30\%-40\%$, through $>90\%$. The $0\%-10\%$ category does not appear because, for each test case, at least one of the classes will have propensity of $10\%$ or greater, so a class with propensity less than $10\%$ will never be selected for any observation. For the ImageNet-style problems that have 1,000 possible labels, as in Figures \ref{fig:count_imagenet} and \ref{fig:count_imagenetv2}, such low propensities are possible, and the lowest grouping is $\le 10\%$. Each of the lines in each graph is a histogram, and the values plotted along the vertical axis show the percent of test cases with propensity scores falling into each of those nine ranges. Thus, when a model has low confidence in its projections---and many of its projections have propensity scores are $\le 20\%$, then the graph will slope downward---with high values on the left and low values on the right---as in the 4x4 and 2x2 panels in Figure \ref{fig:count_logistic}. And when a model has high confidence in its projections---and many of its propensity scores $>90\%$, then the graph will show an upward pattern as in the 32x32 panel in Figure \ref{fig:count_dlares}.

The overall patterns in the histograms in Figure \ref{fig:histogram} are reflective of the acutal differences in forecasting accuracy across the different models and dataset that were shown in Table \ref{tab:performance}. As the first row of graphs in Figure \ref{fig:count_logistic} shows, the logistic classifier tends to be most confident in its projections for the MNIST and Fashion-MNIST datasets in red and green. For the full-sized images, the majority of those cases are classified with greater than 90\% confidence. The confidence is lower for KMNIST and is especially low for the SVHN and CIFAR-10 data---with cases concentrated on the left-hand side of the graph with propensity scores below 30\%. Thus the logistic classifier tends to have higher levels of confidence in its projections for easier MNIST-style problems than for the more difficult problems in the SVHN and CIFAR-10 in blue and violet. Moving from left to right, the level of confidence in the classifications declines steadily. For the rightmost graph, when the downsampling is the greatest, the majority of MNIST, KMNIST, SVHN, and CIFAR-10 cases have propensity scores less than 20\%.

In Figure \ref{fig:count_lenet} in the second row, the results for LeNet-5 show greater levels of confidence for all five datasets---with 70\% or more of the full-size MNIST-style and SVHN images classified with over 90\% confidence. The levels of confidence continue to be relatively low for the more difficult CIFAR-10 problems, with propensities concentrated below 50\% for cases from that dataset. The pattern from left to right is similar as with Figure \ref{fig:count_logistic}, where with increasing pixelation, the concentrations of cases with high propensities fall, and the concentrations with low propensities rise.

In Figure \ref{fig:count_dlares}, the DLA-Simple and ResNet-18 neural network classifiers both classify nearly all of the full-sized SVHN and CIFAR-10 images with greater than 90\% confidence. The degradation in performance as the images are increasingly downsampled appears here as well, with the greater amount of degradation for the more difficult CIFAR-10 problems than for the SVHN cases, and with greater degradation for ResNet-18 than for DLA-Simple.

For the ImageNet-style images in Figures \ref{fig:count_imagenet} and \ref{fig:count_imagenetv2}, the levels of confidence for the full-sized images are generally lower than for the small grayscale and RGB images, a result that is consistent with the greater difficulty of selecting among 1,000 classes relative to 10. For the most effective classifiers, among them ResNetXt101-32x8d, Wide ResNet-101-2, ResNet-101, and DenseNet-201 in solid red, orange, violet, and brown, the fraction of full-sized cases that are classified with greater than 90\% confidence range from 50\% to 60\%. The levels of confidence tend to be somewhat lower for the EfficientNet classifiers and for the earlier and less accurate classifiers, including GoogLeNet, Inception v3, and AlexNet in dashed blue, violet, and brown. As with the other datasets, the concentrations of cases shift from high to low propensity scores for all models with successive levels of downsampling.

Next, the 25 graphs in Figure \ref{fig:correct} show how the models' internal propensity scores co-move with model accuracy. The structure of the graphs is the same as in Figure \ref{fig:histogram} above, but the variable plotted along the vertical axis is now the percentage of cases correctly classified within each propensity score bin. To the extent that the propensity scores are perfect measures of the likelihood that a model's label is correct, the lines should all have constant slopes of one, intersecting the vertical axes on the left at 15\% for Figures \ref{fig:correct_logistic}, \ref{fig:correct_lenet}, and \ref{fig:correct_dlares} and at 5\% for Figures \ref{fig:correct_imagenet} and \ref{fig:correct_imagenetv2} and extending up to 95\% on the right. Cases in which the values on the curves are lower than this indicate that the model are overconfident, and the model does not achieve the level of accuracy predicted by the propensity scores. As noted in Section \ref{downsampling conceptual framework}, however, the validity of the threshold-based approach does not rely upon the propensity scores being unbiased predictors of accuracy---it only requires the relationships to be positive.

For the full-sized images on the left-hand side, the models' levels of correctness generally increase steadily with their propensity scores. Some of the more unusual and jerky patterns arise because they are averages computed over small samples; for instance, the stray $100\%$ red value in the left-hand chart in Figure \ref{fig:correct_lenet} occurs because only LeNet-5 classified only two full-sized test images with $30\%-40\%$ confidence, and its projected classifications were correct in both cases. As we move to the right and greater amounts of downsampling are applied, the explanatory power of the propensity scores declines. For the downsampled images in all five rows, the models tend to be overconfident in their classifications---\emph{i.e.}, they report high propensity scores for many cases in which their actual classification accuracy is low---and for the most extreme amounts of downsampling, the lines on the graph are flat, indicating that the modeled propensity would not be useful for a threshold-based classification approach. For this reason, the smallest images---the 4x4 and 2x2 cases for MNIST-style and small RGB and the 32 and 16 cases for ImageNet and ImageNet-V2---are ignored by the threshold-based approach that is implemented in Section \ref{downsampling results}, and the sequential classifier takes the classifications of the third column of images (7x7, 8x8, and 64) and the corresponding levels of confidence with those classifications as its starting point. Among the neural network classifiers applied to the ImageNet-style images, the Inception v3 model in dashed violet has a particularly high degree of overconfidence, and DenseNet-201 in solid brown also has somewhat more overconfidence than the other models. Alternatively, EfficientNet-b0 in solid green and GoogLeNet in dashed blue tend have accuracy rates at the higher ends of their ranges of propensity scores and may be slightly underconfident about the accuracy of their classifications.

\begin{figure}[ht]
\centering
\begin{subfigure}{.3\columnwidth}
  \centering
  \includegraphics[width=\linewidth]{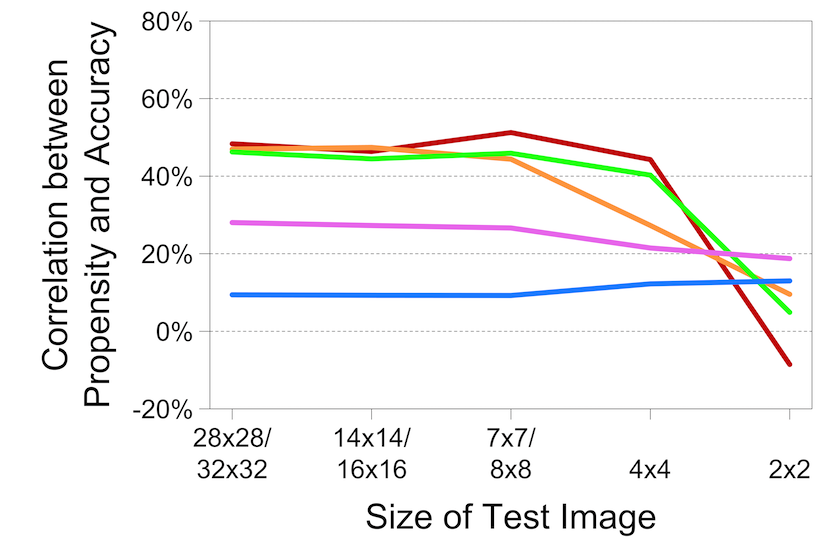}
  \caption{Logistic}
  \label{fig:correl2}
\end{subfigure}
\begin{subfigure}{.3\columnwidth}
  \centering
  \includegraphics[width=\linewidth]{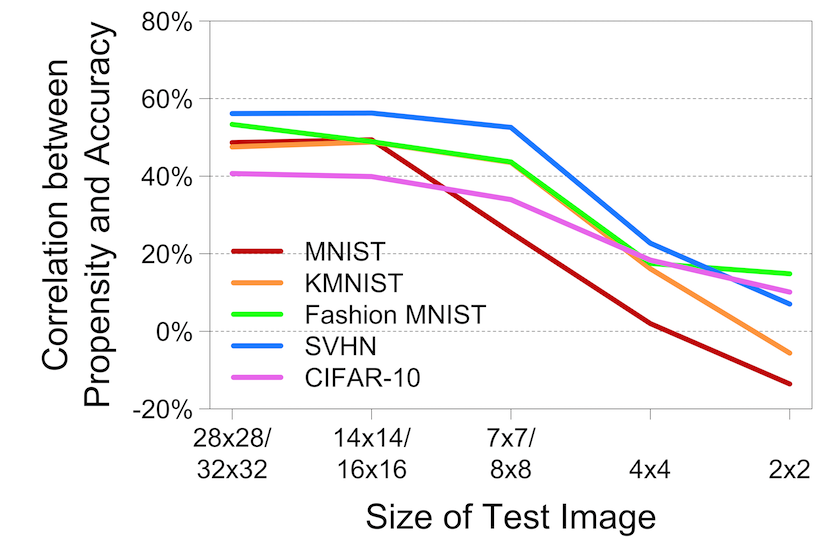}
  \caption{LeNet-5}
  \label{fig:correl1}
\end{subfigure}
\begin{subfigure}{.3\columnwidth}
  \centering
  \includegraphics[width=\linewidth]{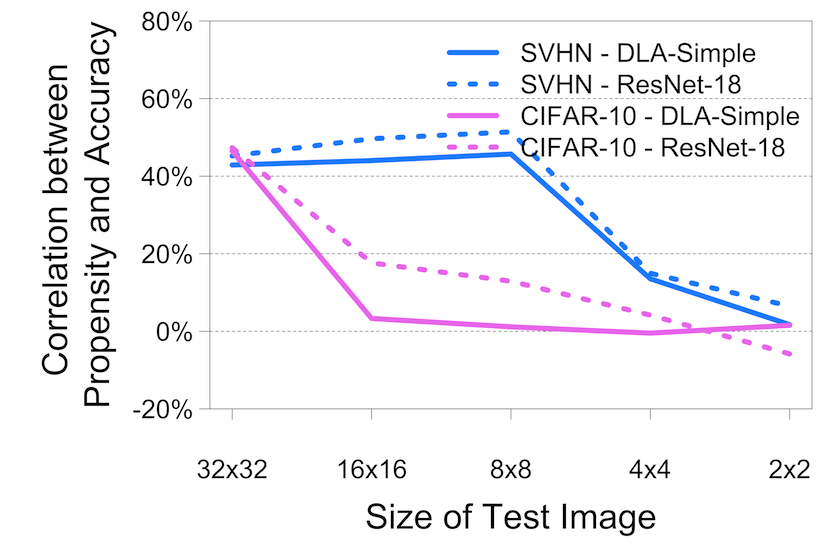}
  \caption{DLA-Simple and ResNet-18}
  \label{fig:correl3}
\end{subfigure}
\begin{subfigure}{.3\columnwidth}
  \centering
  \includegraphics[width=\linewidth]{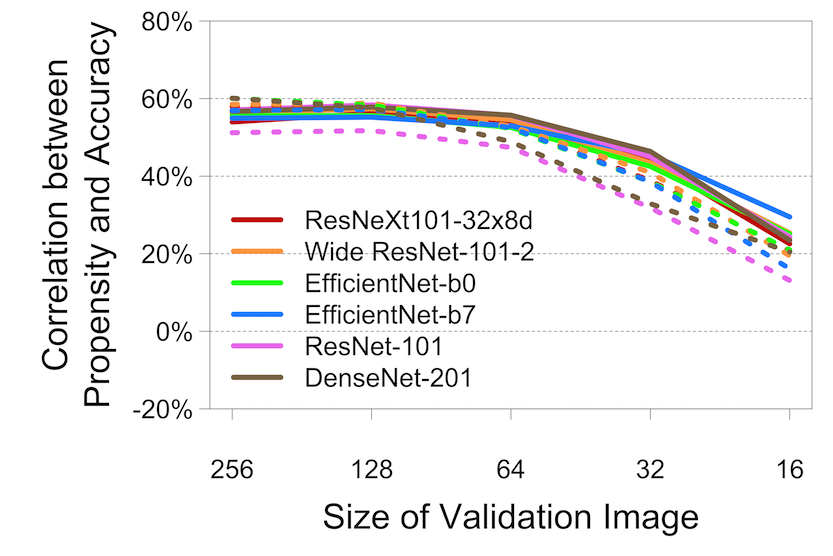}
  \caption{ImageNet}
  \label{fig:correl4}
\end{subfigure}
\begin{subfigure}{.3\columnwidth}
  \centering
  \includegraphics[width=\linewidth]{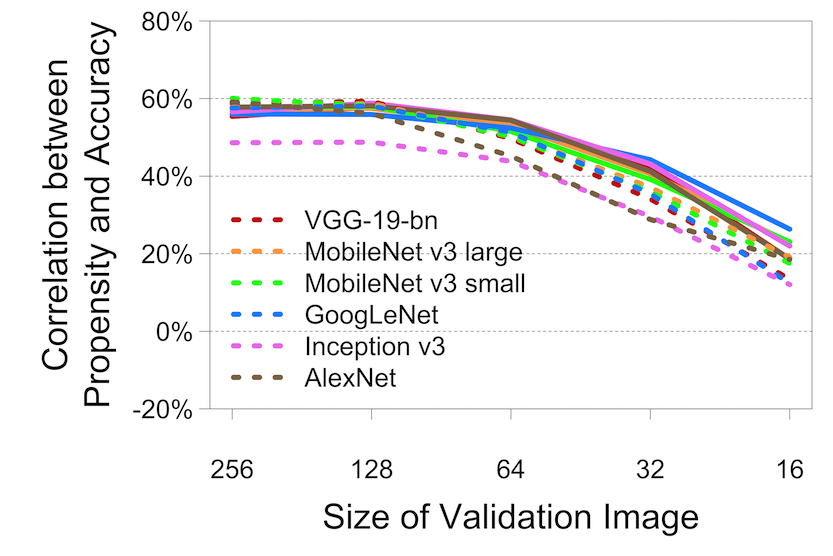}
  \caption{ImageNet-V2}
  \label{fig:correl5}
\end{subfigure}
\hfill 
\begin{minipage}{\linewidth}
\vspace{0.10 in}
\flushleft \small Notes: See notes to Figures \ref{fig:histogram} and Figure \ref{fig:correct}. For each of the datasets, models, and image resolutions considered here, these graphs show the correlations across test cases between the models' levels of confidence in their classifications and an indicator for whether the classifier is correct. Higher values are indicative of higher signal-to-noise ratios and greater amounts of information content coming from the propensity scores. The legend in Figure \ref{fig:correl1} also applies to the graphs in Figure \ref{fig:correl2}, and the legend for Figures \ref{fig:correl4} and \ref{fig:correl5} is split across the two graphs, with all 12 lines appearing in both.
\end{minipage}
\caption{Correlations of Propensities with Correct Classification}
\label{fig:correlation}
\end{figure}

The key findings from Figures \ref{fig:histogram} and \ref{fig:correct} are presented in a consolidated form in Figure \ref{fig:correlation}, which illustrates the correlation between the modeled propensity score and accuracy as it varies by dataset, model, and amount of downsampling. Plotted along the vertical axis is the Pearson correlation coefficient between the propensity score and an indicator for the classification being correct, calculated across all images in the test or validation sample. Plotted along the horizontal axis is the amount of downsampling applied to the test or validation images.

The five graphs in Figure \ref{fig:correlation} show that, for the full-sized images, the propensity scores are reasonably predictive---they are positively correlated with classification accuracy, and the correlations tend to be around $40\%-60\%$, with lower values for the logistic model applied to the small RGB images. These correlations---and the consequent reliability of the propensity scores for use in a threshold-based model---decline substantially as the resolution of the test image declines. In the extreme case of 2x2 images in Figures \ref{fig:correl2}, \ref{fig:correl1}, and \ref{fig:correl3}, these correlations are all under $20\%$ and in many cases negative. The decline is most immediate for the neural network classifications of the CIFAR-10 images in Figure \ref{fig:correl3}. For models applied to the ImageNet and ImageNet-V2 datasets in Figures \ref{fig:correl4} and \ref{fig:correl5}, the impact of downsampling is not quite as severe as for the smaller images; for the images with the smaller side reduced to 16 pixels, the correlation between propensity and correctness ranges from 10\% to 30\%. As was visible in the plots of accuracy versus propensity in Figures \ref{fig:correct_imagenet} and \ref{fig:correct_imagenetv2}, the correlation tends to be lowest for Inception v3 model in dashed violet. In general, the propensities still tend to be informative of classification accuracy when obtained from the middle images with the middle amounts of downsampling (a factor of four in each direction), and so the threshold-based classifier used in this study takes those as a starting point.

\section{Results from Threshold-based Classifiers} \label{downsampling results}

\subsection{Application to Image Resolution} \label{downsampling resolution}

Table \ref{tab:threshold} illustrates the results for the different datasets and models from the sequential classifier based upon image resolution as outlined in Figure \ref{fig:flowchart}, using a variety of different propensity thresholds for case-specific model selection. The structure of the table is similar to that of Table \ref{tab:performance}, but for each dataset and model, two rows of values are shown: the percentage of cases that were correctly classified and the number of bytes of data that were required. The column labeled ``Benchmark'' shows the performance obtained by applying the trained model on the full-sized images, as in the full-sized Test or Validation results column in Table \ref{tab:performance}. In that case, the data cost is equal to the number of bytes of the full-sized image: $28*28 = 784$ for MNIST-style images, $32*32*3 = 3,072$ for the small RGB images, $32*32 = 1,024$ for the grayscale versions of those small RGB images. For the ImageNet and ImageNet-V2 images, which have varying sizes, the median image consumes $500*375*3=562,500$ bytes before scaling.\footnote{For all of the models considered here, the images are rescaled in the benchmark case so that the shorter side has length of 256 pixels. Thus, $562,500*(256/375)^2=262,144$ bytes of information from this median image are used in the benchmark case. For the case in which the full-sized image's shorter side is already less than 256, the number of bytes of the original image is used. Thus, for the ImageNet and ImageNet-V2 cases, average bytes read for the benchmark case is an average over this combination of actual image size for the smaller images and downsampled (to 256) image size for images with both sides longer than 256 pixels.}

In panel A of Table \ref{tab:threshold}, the columns to the right of the benchmark show the results from a sequential model in which the 7x7 test image is used first, the 14x14 image is used if the propensity for that classification falls below the threshold, and if that classification falls below the threshold, the full-sized image is used. Results are shown using thresholds of 80\%, 90\%, 95\%, 97.5\%, 99\%, and 99.9\%. For cases in which the 7x7 image is used, the number of bytes read is only $7*7 = 49$. If the model's level of confidence for that classification falls below the threshold, however, then that image is discarded and a new and larger image must be read. Thus, if the 14x14 image is used, the total number of bytes read is $49 + 14*14 = 245$ for a MNIST-style image. If neither of the downsampled images are used for a given case, then those 245, bytes have already been read, and the classifier must still incur the 784-byte cost of reading the full-sized image. Thus the worst case data cost for the sequential classifier is greater than for the benchmark case, and it rises with the propensity threshold. Within each row, the bolded values show the accuracy and data usage rates for the threshold-based classifier that, among the thresholds presented here, achieves the lowest data usage subject to the constraint that the accuracy is at least 95\% as high as that of the benchmark.

For the small RGB images in panel B, the same setup applies, but the first image examined has dimensionality 8x8x3 and consumes $8*8*3 = 192$ bytes or $8*8 = 64$ for the images that were converted to grayscale for the LeNet-5 classifier. Supposing that the level of confidence falls below the threshold, the 16x16x3 image is examined, so that the total data cost is $192 + 16*16*3 = 960$ or $64 + 16*16 = 320$ for the LeNet-5 case in which the image is converted to grayscale. And if the propensities in both downsampled cases fall below the threshold, then the total data cost is $192 + 960 + 3,072=4,224$ or $64+320+1,024=1,408$ in the LeNet-5 case.

For the ImageNet-style images in panel C, the setup is slightly different because the images have different sizes. The threshold-based classifier first uses images that are downsampled by a rate so that the smaller side has 64 pixels. If the image is already smaller than 64 pixels on both sides, then no downsampling is applied. If the propensity score for the forecast on the 64 image falls below the threshold, then the model generates a projected value for an image that is downsampled by a rate so that the smaller side has 128 pixels, and if that propensity falls below the threshold, then the 256 image (the benchmark) is used. Additionally, a different set of thresholds is used, because propensity scores tend to be lower for the best ImageNet classifiers than for the best classifiers of the smaller images.

\begin{table}
\renewcommand{\arraystretch}{1.1}
\centering
\resizebox{0.45\textwidth}{!}{
\begin{tabular}{cccccccccc}
\multicolumn{10}{c}{Panel A: MNIST-style images} \\
\multirow{2}{*}{Dataset} & \multirow{2}{*}{Model} & \multirow{2}{*}{Outcome} & \multirow{2}{*}{Benchmark} & \multicolumn{6}{c}{Propensity Threshold} \\
 & & & & 80.0\% & 90.0\% & 95.0\% & 97.5\% & 99.0\% & 99.9\% \\
\hline
MNIST & Logistic & \% Correctly Classified & 91.5\% & \textbf{91.5\%} & 91.5\% & 91.5\% & 91.5\% & 91.5\% & 91.5\% \\
 & & Avg. Bytes Read & 784.0 & \textbf{469.3} & 577.1 & 673.2 & 767.6 & 853.2 & 984.5 \\
 & LeNet-5 & \% Correctly Classified & 98.5\% & 70.6\% & 75.3\% & 78.9\% & 82.1\% & 86.2\% & \textbf{94.1\%} \\
 & & Avg. Bytes Read & 784.0 & 102.7 & 130.1 & 153.0 & 176.1 & 209.8 & \textbf{304.6} \\

\hline
KMNIST & Logistic & \% Correctly Classified & 67.9\% & \textbf{67.9\%} & 67.9\% & 67.9\% & 67.9\% & 67.9\% & 67.9\% \\
 & & Avg. Bytes Read & 784 & \textbf{773.9} & 875.2 & 939.1 & 980.0 & 1,007.1 & 1,026.5 \\
 & LeNet-5 & \% Correctly Classified & 92.7\% & 82.4\% & 86.3\% & \textbf{88.8\%} & 90.3\% & 91.3\% & 92.5\% \\
 & & Avg. Bytes Read & 784.0 & 153.7 & 196.4 & \textbf{238.8} & 274.6 & 320.0 & 439.8 \\
\hline
FashionMNIST & Logistic & \% Correctly Classified & 83.8\% & \textbf{83.3\%} & 83.6\% & 83.7\% & 83.7\% & 83.8\% & 83.8\% \\
 & & Avg. Bytes Read & 784.0 & \textbf{532.3} & 640.5 & 726.3 & 795.5 & 867.3 & 973.9 \\
 & LeNet-5 & \% Correctly Classified & 87.7\% & 82.7\% & \textbf{85.0\%} & 85.9\% & 86.8\% & 87.3\% & 87.7\% \\
 & & Avg. Bytes Read & 784.0 & 260.0 & \textbf{342.7} & 409.1 & 473.6 & 547.1 & 701.3 \\
\\
\multicolumn{10}{c}{Panel B: Small RGB images} \\
\multirow{2}{*}{Dataset} & \multirow{2}{*}{Model} & \multirow{2}{*}{Outcome} & \multirow{2}{*}{Benchmark} & \multicolumn{6}{c}{Propensity Threshold} \\
 & & & & 80.0\% & 90.0\% & 95.0\% & 97.5\% & 99.0\% & 99.9\% \\
\hline
SVHN  & Logistic & \% Correctly Classified & 25.2\% & \textbf{25.2\%} & 25.2\% & 25.2\% & 25.2\% & 25.2\% & 25.2\% \\
 & & Avg. Bytes Read & 3,072.0 & \textbf{4,030.1} & 4,031.6 & 4,031.9 & 4,032.0 & 4,032.0 & 4,032.0 \\
 & LeNet-5 (Grayscale) & \% Correctly Classified & 83.1\% & \textbf{81.3\%} & 82.3\% & 82.7\% & 82.9\% & 83.0\% & 83.1\% \\
 & & Avg. Bytes Read & 1,024.0 & \textbf{378.3} & 477.6 & 565.0 & 647.2 & 752.1 & 1,026.4 \\
 & DLA-Simple & \% Correctly Classified & 97.3\% & 89.4\% & 91.0\% & 92.3\% & \textbf{93.4\%} & 94.5\% & 97.0\% \\
 & & Avg. Bytes Read & 3,072.0 & 319.3 & 369.0 & 422.7 & \textbf{474.9} & 555.2 & 2,596.5 \\
 & ResNet-18 & \% Correctly Classified & 97.2\% & 90.7\% & \textbf{92.5\%} & 93.8\% & 94.8\% & 95.6\% & 97.1\% \\
 & & Avg. Bytes Read & 3,072.0 & 380.3 & \textbf{449.5} & 518.3 & 585.4 & 694.5 & 2,088.6 \\
\\
\hline
CIFAR-10 & Logistic & \% Correctly Classified & 41.0\% & \textbf{41.0\%} & 41.0\% & 41.0\% & 41.0\% & 41.0\% & 41.0\% \\
 & & Avg. Bytes Read & 3,072.0 & \textbf{3,924.4} & 4,001.4 & 4,021.2 & 4,028.7 & 4,031.6 & 4,032.0 \\
 & LeNet-5 (Grayscale) & \% Correctly Classified & 49.7\% & \textbf{49.6\%} & 49.7\% & 49.7\% & 49.7\% & 49.7\% & 49.7\% \\
 & & Avg. Bytes Read & 1,024.0 & \textbf{1,213.4} & 1,286.9 & 1,323.0 & 1,337.7 & 1,343.3 & 1,344.0 \\
 & DLA-Simple & \% Correctly Classified & 94.9\% & 36.6\% & 45.0\% & 51.4\% & 57.2\% & 64.1\% & 81.7\% \\
 & & Avg. Bytes Read & 3,072.0 & 1,156.9 & 1,539.5 & 1,863.3 & 2,155.2 & 2,502.3 & 3,346.7 \\
 & ResNet-18 & \% Correctly Classified & 95.4\% & 64.0\% & 75.7\% & 82.4\% & 87.1\% & \textbf{91.3\%} & 95.0\% \\
 & & Avg. Bytes Read & 3,072.0 & 2,338.6 & 2,898.7 & 3,248.6 & 3,492.4 & \textbf{3,728.5} & 3,980.7 \\
\\
\end{tabular}}
\resizebox{0.50\textwidth}{!}{
\begin{tabular}{ccccccccccc}
\multicolumn{11}{c}{Panel C: Larger RGB images} \\
\multirow{2}{*}{Dataset} & \multirow{2}{*}{Model} & \multirow{2}{*}{Outcome} & \multirow{2}{*}{Benchmark} & \multicolumn{7}{c}{Propensity Threshold} \\
 & & & & 50.0\% & 60.0\% & 70.0\% & 80.0\% & 90.0\% & 95.0\% & 97.5\% \\
\hline
ImageNet & ResNeXt-101-32x8d & \% Correctly Classified & 79.3\% & 65.1\% & 68.6\% & 71.4\% & 74.0\% & \textbf{76.3\%} & 77.5\% & 78.3\% \\
 & & Avg. Bytes Read & 258,359 & 59,965 & 78,620 & 98,632 & 121,885 & \textbf{153,494} & 179,431 & 201,171 \\
 & Wide ResNet 101-2 & \% Correctly Classified & 78.8\% & 66.2\% & 69.6\% & 72.3\% & 74.6\% & \textbf{76.6\%} & 77.6\% & 78.2\% \\
 & & Avg. Bytes Read & 258,359 & 78,132 & 99,575 & 122,513 & 147,889 & \textbf{182,121} & 208,444 & 229,695 \\
 & EfficientNet-b0 & \% Correctly Classified & 77.7\% & 71.6\% & \textbf{73.9\%} & 75.6\% & 76.7\% & 77.4\% & 77.6\% & 77.7\% \\
 & & Avg. Bytes Read & 258,359 & 129,588 & \textbf{160,335} & 191,144 & 226,501 & 274,769 & 309,318 & 327,446 \\
 & EfficientNet-b7 & \% Correctly Classified & 73.9\% & 64.1\% & 67.1\% & 69.5\% & \textbf{71.6\%} & 73.3\% & 73.8\% & 73.9\% \\
 & & Avg. Bytes Read & 258,359 & 93,812 & 121,404 & 152,876 & \textbf{194,605} & 269,439 & 323,438 & 337,693 \\
 & ResNet-101 & \% Correctly Classified & 77.4\% & 66.8\% & 70.1\% & 72.6\% & \textbf{74.4\%} & 76.0\% & 76.7\% & 77.0\% \\
 & & Avg. Bytes Read & 258,359 & 82,065 & 106,177 & 131,472 & \textbf{158,702} & 193,727 & 219,958 & 241,825 \\
 & DenseNet-201 & \% Correctly Classified & 76.9\% & 67.5\% & 70.4\% & 72.6\% & \textbf{74.4\%} & 75.8\% & 76.4\% & 76.6\% \\
 & & Avg. Bytes Read & 258,359 & 92,210 & 117,147 & 143,000 & \textbf{170,933} & 206,610 & 232,754 & 253,538 \\
 & VGG-19-bn & \% Correctly Classified & 74.2\% & 63.1\% & 66.8\% & 69.5\% & \textbf{71.4\%} & 73.0\% & 73.7\% & 73.9\% \\
 & & Avg. Bytes Read & 258,359 & 122,193 & 149,932 & 176,781 & \textbf{204,649} & 237,472 & 260,725 & 277,028 \\
 & MobileNet v3 large & \% Correctly Classified & 74.1\% & 62.2\% & 66.0\% & 68.6\% & \textbf{70.8\%} & 72.6\% & 73.3\% & 73.7\% \\
 & & Avg. Bytes Read & 258,359 & 96,389 & 122,561 & 148,476 & \textbf{176,535} & 212,297 & 237,790 & 257,654 \\
 & MobileNet v3 small & \% Correctly Classified & 67.7\% & 59.1\% & 62.0\% & 64.2\% & \textbf{65.8\%} & 67.0\% & 67.4\% & 67.5\% \\
 & & Avg. Bytes Read & 258,359 & 147,646 & 176,700 & 204,872 & \textbf{232,694} & 263,742 & 283,429 & 297,677 \\
 & GoogLeNet & \% Correctly Classified & 69.8\% & 65.8\% & \textbf{67.5\%} & 68.5\% & 69.2\% & 69.6\% & 69.7\% & 69.8\% \\
 & & Avg. Bytes Read & 258,359 & 177,552 & \textbf{205,174} & 230,526 & 256,355 & 285,621 & 304,662 & 318,017 \\
 & Inception v3 & \% Correctly Classified & 69.5\% & 51.1\% & 54.4\% & 57.1\% & 60.0\% & 63.2\% & 65.3\% & \textbf{66.6\%} \\
 & & Avg. Bytes Read & 258,359 & 47,736 & 64,019 & 82,111 & 104,137 & 134,737 & 159,107 & \textbf{180,096} \\
 & AlexNet & \% Correctly Classified & 56.6\% & 52.2\% & \textbf{54.1\%} & 55.2\% & 55.9\% & 56.3\% & 56.5\% & 56.5\% \\
 & & Avg. Bytes Read & 258,359 & 220,372 & \textbf{245,394} & 266,270 & 285,461 & 303,828 & 315,291 & 322,462 \\
 \\
 \hline
ImageNet V2 & ResNeXt-101-32x8d & \% Correctly Classified & 67.5\% & 55.5\% & 58.4\% & 60.9\% & 63.3\% & \textbf{65.4\%} & 66.5\% & 67.0\% \\
(Matched & & Avg. Bytes Read & 269,906 & 85,099 & 107,486 & 133,312 & 160,845 & \textbf{195,331} & 220,972 & 241,689 \\
Frequency) & Wide ResNet 101-2 & \% Correctly Classified & 66.5\% & 55.4\% & 58.7\% & 61.3\% & \textbf{63.4\%} & 65.0\% & 65.8\% & 66.3\% \\
 & & Avg. Bytes Read & 269,906 & 108,593 & 136,139 & 162,340 & \textbf{190,290} & 224,131 & 249,503 & 268,719 \\
 & EfficientNet-b0 & \% Correctly Classified & 65.7\% & 60.4\% & \textbf{62.6\%} & 64.1\% & 65.0\% & 65.6\% & 65.7\% & 65.7\% \\
 & & Avg. Bytes Read & 269,906 & 168,668 & \textbf{200,611} & 232,497 & 265,243 & 306,650 & 333,474 & 347,151 \\
 & EfficientNet-b7 & \% Correctly Classified & 61.8\% & 53.3\% & 56.0\% & 58.3\% & \textbf{60.2\%} & 61.3\% & 61.7\% & 61.8\% \\
 & & Avg. Bytes Read & 269,906 & 123,454 & 155,254 & 188,696 & \textbf{230,805} & 298,188 & 342,144 & 354,069 \\
 & ResNet-101 & \% Correctly Classified & 65.6\% & 56.2\% & 59.4\% & 61.5\% & \textbf{63.1\%} & 64.6\% & 65.1\% & 65.4\% \\
 & & Avg. Bytes Read & 269,906 & 112,046 & 140,758 & 169,009 & \textbf{198,677} & 233,301 & 259,317 & 278,642 \\
 & DenseNet-201 & \% Correctly Classified & 64.7\% & 56.8\% & 59.2\% & 61.4\% & \textbf{62.6\%} & 64.0\% & 64.4\% & 64.6\% \\
 & & Avg. Bytes Read & 269,906 & 125,033 & 153,994 & 182,175 & \textbf{212,298} & 247,994 & 271,937 & 289,848 \\
 & VGG-19-bn & \% Correctly Classified & 61.9\% & 52.6\% & 55.8\% & 57.9\% & \textbf{59.3\%} & 61.0\% & 61.5\% & 61.7\% \\
 & & Avg. Bytes Read & 269,906 & 156,944 & 187,918 & 217,135 & \textbf{244,350} & 275,148 & 295,476 & 308,850 \\
 & MobileNet v3 large & \% Correctly Classified & 60.5\% & 51.2\% & 54.0\% & 56.5\% & \textbf{58.2\%} & 59.4\% & 60.0\% & 60.2\% \\
 & & Avg. Bytes Read & 269,906 & 132,055 & 160,878 & 190,816 & \textbf{219,579} & 253,137 & 276,805 & 294,479 \\
 & MobileNet v3 small & \% Correctly Classified & 54.7\% & 48.2\% & 50.4\% & \textbf{52.2\%} & 53.5\% & 54.2\% & 54.5\% & 54.7\% \\
 & & Avg. Bytes Read & 269,906 & 183,410 & 215,403 & \textbf{245,067} & 269,642 & 296,336 & 313,218 & 324,510 \\
 & GoogLeNet & \% Correctly Classified & 57.9\% & \textbf{55.1\%} & 56.3\% & 56.9\% & 57.4\% & 57.7\% & 57.9\% & 57.9\% \\
 & & Avg. Bytes Read & 269,906 & \textbf{216,870} & 243,166 & 267,048 & 290,038 & 314,825 & 330,020 & 340,618 \\
 & Inception v3 & \% Correctly Classified & 57.6\% & 40.9\% & 43.8\% & 46.5\% & 49.1\% & 52.2\% & 54.1\% & \textbf{55.5\%} \\
 & & Avg. Bytes Read & 269,906 & 62,799 & 82,504 & 106,178 & 132,491 & 168,193 & 195,594 & \textbf{216,531} \\
 & AlexNet & \% Correctly Classified & 43.5\% & 40.4\% & \textbf{41.6\%} & 42.4\% & 42.9\% & 43.3\% & 43.5\% & 43.5\% \\
 & & Avg. Bytes Read & 269,906 & 254,410 & \textbf{278,403} & 296,812 & 313,530 & 328,939 & 337,665 & 343,088 \\
\hline
\end{tabular}}
\resizebox{0.80\textwidth}{!}{
\begin{tabular}{p{\linewidth}}
\vspace{-0.1 in}
\tiny Notes: See notes to Table \ref{tab:performance}. This table illustrates the performance on different datasets and models at different thresholds of the threshold-based classification model described in Figure \ref{fig:flowchart1}. The benchmark model applies the trained model on a full-resolution test or validation case; the data cost is 28x28 = 784 bytes per MNIST-style image, 32x32x3 = 3,072 bytes per small RGB image, and 32x32 = 1,024 bytes per 32x32 image that is reduced to grayscale. For ImageNet and ImageNet-V2 images, the image sizes and consequent data costs vary across cases in the validation set. For the median case, the data cost is 262,144 bytes. The threshold-based approach begins by applying the classifier on a test or validation case that is downsampled by a rate of 4 in each dimension---so that the amount of bytes read is one sixteenth of the amount in the benchmark case. If the modeled propensity exceeds the threshold, that classification is used. If not, the classifier is applied on a test case that is downsampled by a rate of 2 in each dimension. If the new modeled propensity exceeds the threshold, then that classification is used, otherwise the classifier is applied on the full-resolution test image and that classification is used. The \textbf{bolded} values in each row indicate, among the threshold-based classifiers presented here, the one that minimizes data use subject to the constraint that its performance is at least 95\% as large as the benchmark's.
\end{tabular}}
\caption{Test Accuracy and Data Cost of Sequential Image Resolution-Based Classifier by Dataset, Model, and Propensity Threshold} 
\label{tab:threshold}
\end{table}

\begin{figure}
\centering
\begin{subfigure}{0.48\textwidth}
  \centering
  \includegraphics[width=\textwidth]{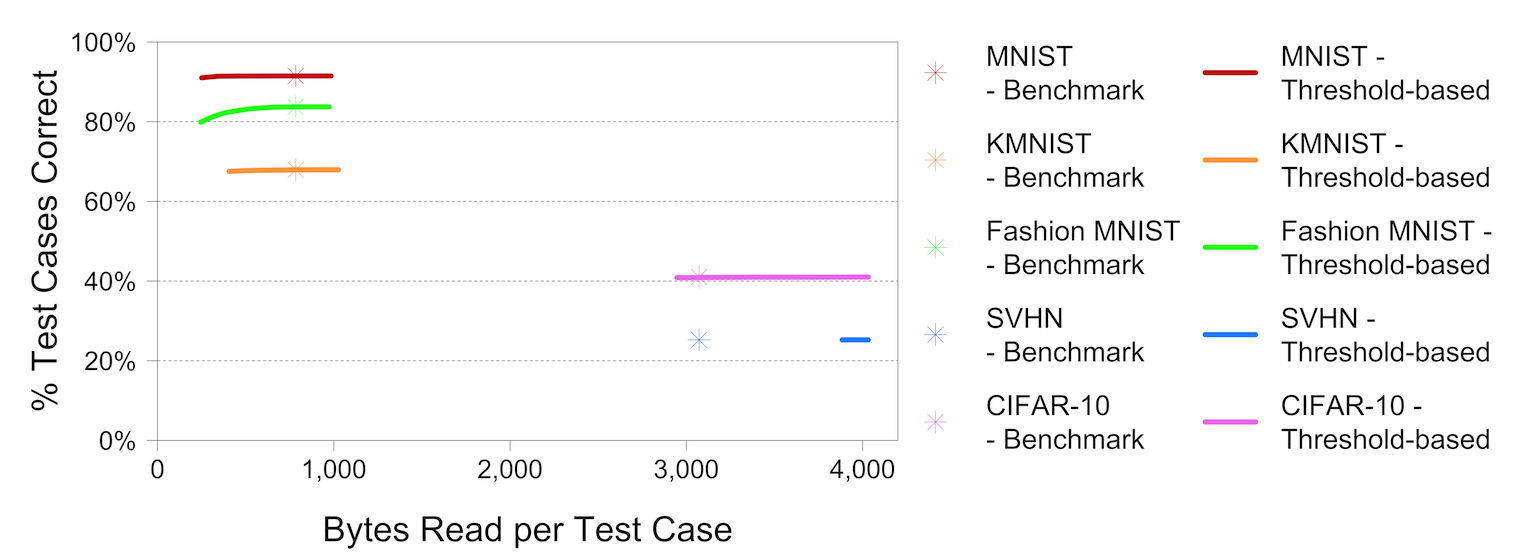}
  \caption{Logistic}
  \label{fig:ppf2}
\end{subfigure}
\begin{subfigure}{0.48\textwidth}
  \centering
  \includegraphics[width=\textwidth]{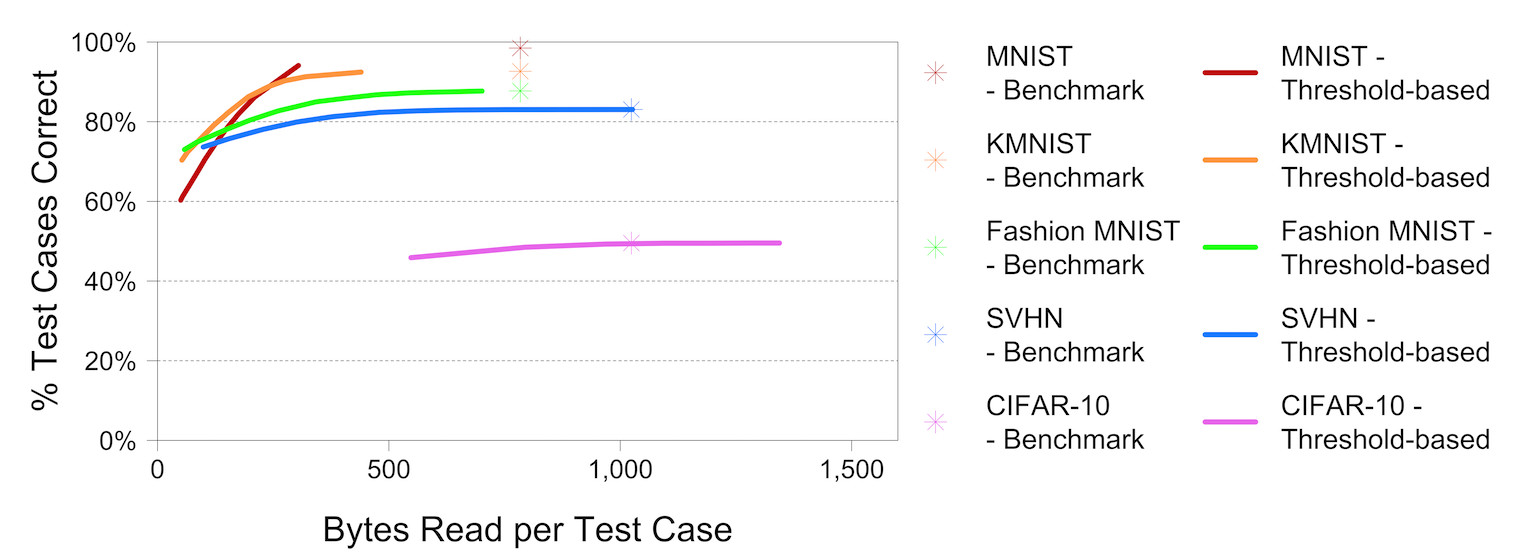}
  \caption{LeNet-5}
  \label{fig:ppf1}
\end{subfigure}
\begin{subfigure}{0.48\textwidth}
  \centering
  \includegraphics[width=\textwidth]{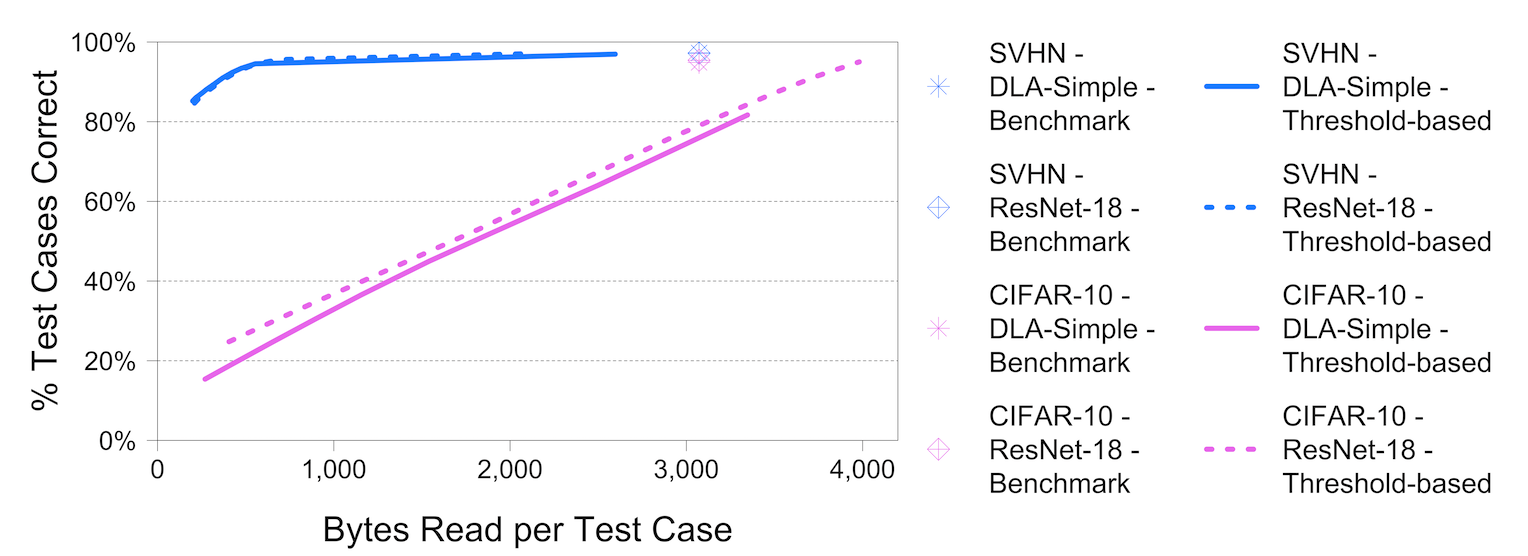}
  \caption{DLA-Simple and ResNet-18}
  \label{fig:ppf3}
\end{subfigure}
\begin{subfigure}{0.48\textwidth}
  \centering
  \includegraphics[width=\textwidth]{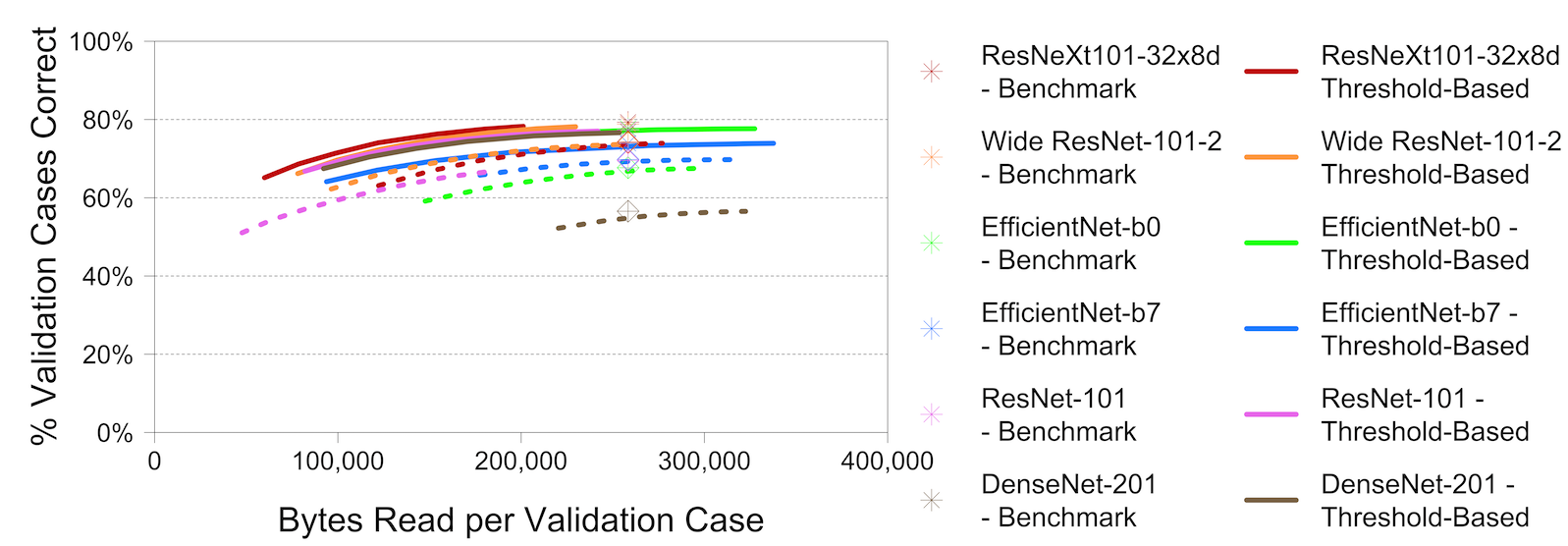}
  \caption{ImageNet}
  \label{fig:ppf4}
\end{subfigure}
\begin{subfigure}{0.48\textwidth}
  \centering
  \includegraphics[width=\textwidth]{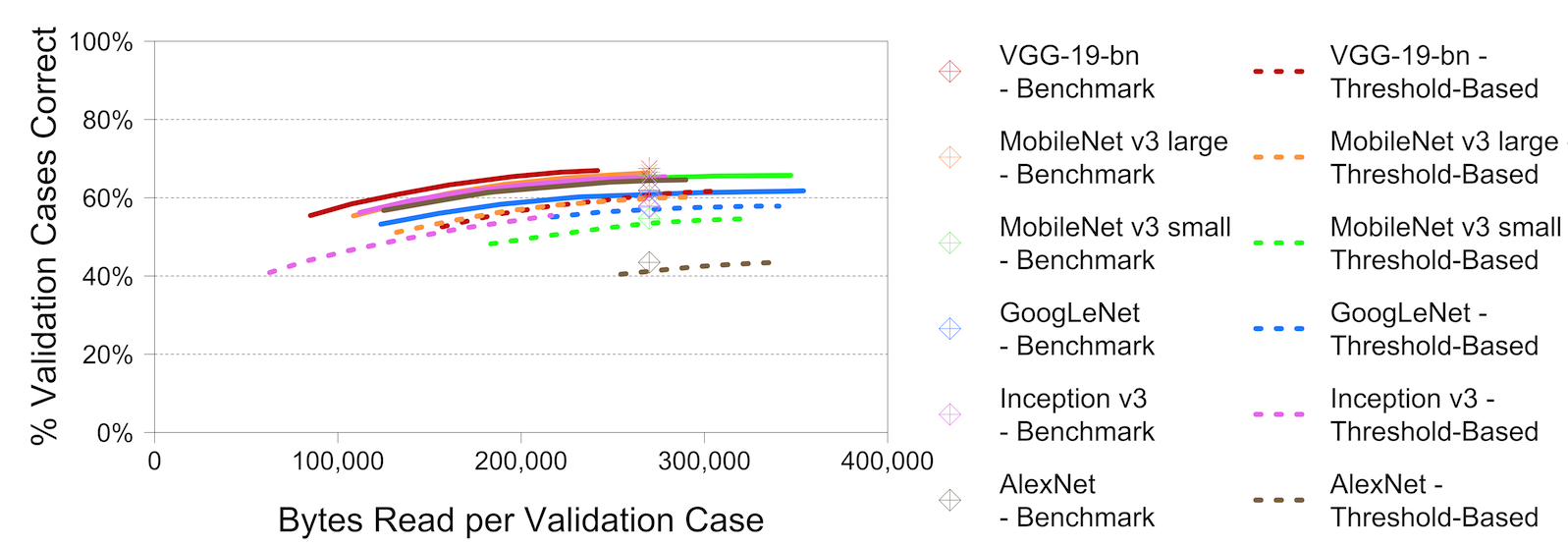}
  \caption{ImageNet-V2}
  \label{fig:ppf5}
\end{subfigure}
\begin{minipage}{\linewidth}
\vspace{0.10 in}
\flushleft \small Notes: See notes to Table \ref{tab:threshold} and Figures \ref{fig:histogram} to \ref{fig:correlation}. These three graphs illustrate the achievable combinations of accuracy and data usage obtained by the ensemble classifiers applied to the datasets and models examined in Table \ref{tab:threshold}; the results are the same as in Table \ref{tab:threshold} but are presented graphically to make the tradeoff between performance and cost more apparent. The legend for Figures \ref{fig:ppf4} and \ref{fig:ppf5} is split across those two figures, and each of those two graphs shows each list benchmark and threshold-based results for all twelve of the neural networks considered for those datasets.
\end{minipage}
\caption{Accuracy versus Data Usage of Threshold-based Estimators}
\label{fig:ppf}
\end{figure}

As the results from panels A and B of Table \ref{tab:threshold} show, for the logistic model, the results vary across the datasets considered. For the MNIST and Fashion MNIST datasets, the sequential classifier with an 80\% threshold achieves roughly the same accuracy as the benchmark but with 32.1\% to 40.1\% less data usage. For the KMNIST, however, the impacts are negligible, and for the SVHN and CIFAR-10, the propensities from the downsampled images consistently fall below the thresholds, so that accuracy is unaffected but data cost is increased.

For the LeNet-5 model, for four out of five datasets on which the model is applied, there is some threshold at which the sequential classifier achieves substantial data savings relative to the benchmark with minimal reduction in classification accuracy. For the MNIST data, for example, when the 99.9\% threshold is used, the LeNet-5 model achieves accuracy of 94.1\% while consuming only 304.6 bytes of data---less than half of what is required in the benchmark case. The sequential approach with the LeNet-5 model performs similarly well with the KMNIST, Fashion MNIST, and SVHN datasets. For KMNIST, for instance, using the 95.0\% threshold produces classification accuracy of 88.8\% while consuming only 238.8 bytes. For Fashion MNIST, using the 90.0\% threshold produces accuracy of 85.0\% while consuming only 342.7 bytes, and for SVHN, using the 90.0\% threshold produces accuracy of 81.3\% while consuming only 378.3 bytes relative to 1,024 bytes in the benchmark application. In all four of these cases, the degradation in performance is less than 5\%, and the data cost is reduced by 56.3\% to 69.5\%. For the CIFAR-10, however, the threshold-based sequential LeNet-5 classifier only increases data usage relative to the benchmark.\footnote{By comparison, the resolution-adaptive network of \cite{Yang2020b}, which is designed to reduce computational intensity rather than bandwidth or data, reduces Floating Point Operations per Second (FLOPS) by 20\% to 65\% relative to well-known classifiers, also with minimal loss in accuracy.}

With the DLA-Simple and ResNet-18 classifiers in panel B, using the threshold-based approach on the SVHN substantially reduces data usage. The DLA-Simple with a 97.5\% threshold achieves 93.4\% accuracy with an 84.5\% reduction in data usage, and the ResNet-18 with a 90\% threshold achieves 90.7\% accuracy with an 85.4\% reduction in data usage. For the CIFAR-10, the results continue to disappoint; the threshold-based classifier consistently reduces performance and in many cases increases the number of bytes read.

For the ImageNet and ImageNet-V2 images in panel C of Table \ref{tab:threshold}, the performance of the threshold-based estimator varies widely across the 12 neural network classifiers considered here. Of the 24 model-data combinations, there are 23 for which a threshold-based approach reduces data usage with a less than 5\% reduction in accuracy. The one exception is AlexNet applied to the ImageNet-V2 data. As the bottom row of panel C shows, only the 50\% threshold estimator reduces data usage in that case but with a 7.0\% reduction in accuracy, from 43.5\% to 40.4\%. For the benchmark case when applied to the ImageNet validation sample, the 12 models considered here achieve an average accuracy of 73.0\% with an average data usage of 258,359 bytes. Among the threshold-based estimators indicated in bold, the average accuracy is 70.3\% with an average data usage of 188,728 bytes. Hence, across these 12 models, the threshold-based estimator reduces data usage by 27.0\% at the expense of a 3.69\% reduction in accuracy. The reductions in data usage are greatest for ResNeXt-101-32x8d (a 40.6\% savings), ResNet-101 (a 38.6\% savings), EfficientNet-b7 (a 37.9\% savings), and DenseNet-201 (a 33.8\% savings). When applied to the ImageNet-V2 validation sample, the average benchmark has accuracy of 60.7\% and data usage of 269,906 bytes. Among the threshold-based estimators shown in bold for this dataset, the average accuracy is 58.3\%, and the average data usage is 220,734 bytes---so that the average estimator reduces data usage by 18.2\% at a cost of a 3.95\% reduction in accuracy. The models for which the threshold-based estimator produces the greatest reduction in data usage are Wide ResNet 101-2 with a savings of 29.5\%, ResNeXt 101-32x8d with a savings of 27.6\%, ResNet-101 with a savings of 26.4\%, and EfficientNet-b0 with a savings of 25.7\%.

The five charts in Figure \ref{fig:ppf} illustrate the tradeoffs between performance and data requirements that are implied by the results in Table \ref{tab:threshold}. For each of the classifiers whose results are shown in that table, the classification accuracy is plotted along the vertical axis of one of the three figures, and the amount of data used for the average test case is plotted along the horizontal axis. The ideal classifier is represented by the combination of values in the top left of each graph, with 100\% accuracy and zero bytes read. Each colored point or line shows an achievable combination of these two objectives. The asterisks in all three charts and the diamonds in Figure \ref{fig:ppf3} present the performance and resource intensity of the benchmark classifiers from Table \ref{tab:threshold}. The lines illustrate the combinations obtained from the sequential classifiers with propensity thresholds ranging from 80\% to 99.9\% as in the table. As with Figure \ref{fig:correlation}, the first chart shows results for the LeNet-5, the second for the logistic regression approach, the third for DLA-Simple and ResNet-18, and the fourth and fifth are for the twelve networks estimated on the ImageNet and ImageNet-V2 datasets.

The patterns shown in the graphs reflect the same tradeoffs as appear in the table. For the logistic regression model in Figure \ref{fig:ppf2}, cost reductions are achievable for two of the five datasets. For the LeNet-5 model in Figure \ref{fig:ppf1}, using the sequential classifier enables movement to the left (a reduction in data usage) with minimal downward movement (reduction in accuracy) with four out of five of the datasets. For the DLA-Simple and ResNet-18 classifiers in \ref{fig:ppf3}, substantial savings are achieved on the SVHN, but for the CIFAR-10, the threshold-based classifiers that are able to reduce data usage entail too large of a loss in performance to be viable. Among the ImageNet and ImageNet-V2 datasets in Figures \ref{fig:ppf4} and \ref{fig:ppf5}, all of the models show a downward slope for both datasets, indicating real tradeoffs. But for the most effective classifiers, the solid lines at the top of both graphs achieve close to the levels of performance shown of the benchmarks (indicated by the stars and diamonds) but with noticeable savings in data use.

\subsection{Application to Computational Complexity} \label{downsampling complexity}

Table \ref{tab:gmacs_threshold} presents the main results from the second application in this study. Classification accuracy is shown along with two measures of computational complexity: MMACs and miliseconds of time per case on a MacBook Pro. The first column of results shows the top-performing benchmark, and the second column shows the ``Max Propensity'' result that requires running all of the candidates. The remaining columns show the results of the threshold-based estimators.

\begin{table}
\renewcommand{\arraystretch}{1.1}
\centering
\resizebox{0.70\textwidth}{!}{
\begin{tabular}{cccccccccc}
\multicolumn{10}{c}{Panel A: MNIST-style images} \\
\multirow{2}{*}{Dataset} & \multirow{2}{*}{Outcome} & Top Bench- & Max Prop- & \multicolumn{6}{c}{Propensity Threshold} \\
 & & mark & ensity & 70.0\% & 80.0\% & 90.0\% & 95.0\% & 97.5\% & 99.9\% \\
\hline
MNIST & \% Correctly Classified & 98.5\% & 98.4\% & \textbf{96.5\%} & 97.2\% & 97.9\% & 98.1\% & 98.3\% & 98.3\% \\
 & MMACs & 0.429 & 0.429 & \textbf{0.077} & 0.102 & 0.146 & 0.188 & 0.230 & 0.284 \\
 & Miliseconds & 0.10 & 0.10 & \textbf{0.02} & 0.02 & 0.03 & 0.04 & 0.05 & 0.07 \\
KMNIST & \% Correctly Classified & 92.7\% & 92.4\% & \textbf{88.4\%} & 90.1\% & 91.7\% & 92.1\% & 92.2\% & 92.4\% \\
 & MMACs & 0.429 & 0.429 & \textbf{0.225} & 0.270 & 0.326 & 0.364 & 0.389 & 0.412 \\
 & Miliseconds & 0.10 & 0.10 & \textbf{0.05} & 0.06 & 0.08 & 0.08 & 0.09 & 0.10 \\
FashionMNIST & \% Correctly Classified & 87.7\% & 88.1\% & \textbf{87.3\%} & 87.7\% & 87.9\% & 88.0\% & 88.1\% & 88.1\% \\
 & MMACs & 0.429 & 0.429 & \textbf{0.119} & 0.157 & 0.204 & 0.242 & 0.272 & 0.309 \\
 & Miliseconds & 0.10 & 0.10 & \textbf{0.03} & 0.04 & 0.05 & 0.06 & 0.06 & 0.07 \\
\\
\multicolumn{10}{c}{Panel B: Small RGB images} \\
\multirow{2}{*}{Dataset} & \multirow{2}{*}{Outcome} & Top Bench- & Max Prop- & \multicolumn{6}{c}{Propensity Threshold} \\
 & & mark & ensity & 70.0\% & 80.0\% & 90.0\% & 95.0\% & 97.5\% & 99.9\% \\
\hline
SVHN & \% Correctly Classified & 97.3\% & 97.4\% & 90.7\% & \textbf{92.7\%} & 94.7\% & 96.0\% & 96.6\% & 97.0\% \\
 & GMACs & 0.916 & 1.473 & 0.087 & \textbf{0.120} & 0.169 & 0.216 & 0.216 & 0.326 \\
 & Miliseconds & 5.8 & 10.7 & 0.8 & \textbf{1.1} & 1.5 & 1.9 & 2.3 & 2.8 \\
CIFAR-10 & \% Correctly Classified & 95.4\% & 95.8\% & 88.5\% & \textbf{92.2\%} & 94.9\% & 95.5\% & 95.7\% & 95.8\% \\
 & GMACs & 0.557 & 1.473 & 0.409 & \textbf{0.482} & 0.562 & 0.609 & 0.642 & 0.679 \\
 & Miliseconds & 4.8 & 10.7 & 3.6 & \textbf{4.2} & 4.8 & 5.2 & 5.4 & 5.7 \\
\\
\end{tabular}}
\resizebox{0.80\textwidth}{!}{
\begin{tabular}{ccccccccccccc}
\multicolumn{13}{c}{Panel C: Larger RGB images} \\
\multirow{2}{*}{Dataset} & \multirow{2}{*}{Models} & \multirow{2}{*}{Outcome} & Top Bench- & Max Prop- & \multicolumn{8}{c}{Propensity Threshold} \\
 & & & mark & ensity & 50.0\% & 60.0\% & 70.0\% & 80.0\% & 90.0\% & 95.0\% & 97.5\% & 99.9\% \\
\hline
ImageNet & All & \% Correctly Classified & 79.3\% & 79.8\% & 73.9\% & 75.3\% & \textbf{76.5\%} & 77.3\% & 78.1\% & 78.4\% & 78.6\% & 78.7\% \\
 & & GMACs & 16.51 & 82.29 & 1.10 & 2.18 & \textbf{4.19} & 7.80 & 14.28 & 20.35 & 25.63 & 31.92 \\
& & Miliseconds & 67.4 & 580.0 & 25.4 & 35.9 & \textbf{52.9} & 80.4 & 126.5 & 168.7 & 205.4 & 248.8 \\
& Set 1 & \% Correctly Classified & 79.3\% & 81.0\% & \textbf{77.4\%} & 78.5\% & 79.3\% & 80.0\% & 80.4\% & 80.5\% & 80.5\% & 80.6\% \\
 & & GMACs & 16.51 & 52.18 & \textbf{2.41} & 4.15 & 6.58 & 10.01 & 14.91 & 19.06 & 22.70 & 27.04 \\
& & Miliseconds & 67.4 & 244.3 & \textbf{28.8} & 37.9 & 49.6 & 65.3 & 87.1 & 105.5 & 121.5 & 140.5 \\
& Set 2 & \% Correctly Classified & 79.3\% & 80.3\% & \textbf{77.3\%} & 78.5\% & 79.2\% & 79.7\% & 80.0\% & 80.1\% & 80.1\% & 80.1\% \\
 & & GMACs & 16.51 & 21.51 & \textbf{1.79} & 2.83 & 4.10 & 5.73 & 7.89 & 9.71 & 11.26 & 13.08 \\
& & Miliseconds & 67.4 & 143.5 & \textbf{26.9} & 33.8 & 41.7 & 51.5 & 64.2 & 75.0 & 84.1 & 94.7 \\
\\
\hline
ImageNet-V2 & All & \% Correctly Classified & 67.5\% & 68.1\% & 61.5\% & 63.0\% & 64.1\% & \textbf{65.3\%} & 66.2\% & 66.7\% & 67.0\% & 67.2\% \\
 & & GMACs & 16.51 & 82.29 & 2.11 & 4.24 & 7.40 & \textbf{12.59} & 21.21 & 28.56 & 34.75 & 41.13 \\
& & Miliseconds & 67.4 & 580.0 & 36.4 & 55.0 & 80.0 & \textbf{118.0} & 177.6 & 227.3 & 268.9 & 311.7 \\
& Set 1 & \% Correctly Classified & 67.5\% & 69.4\% & \textbf{65.2\%} & 66.4\% & 67.5\% & 68.2\% & 68.7\% & 68.9\% & 68.9\% & 69.0\% \\
 & & GMACs & 16.51 & 52.18 & \textbf{4.38} & 6.97 & 10.42 & 14.83 & 20.63 & 25.29 & 29.16 & 33.14 \\
& & Miliseconds & 67.4 & 244.3 & \textbf{39.7} & 52.7 & 68.8 & 88.3 & 113.3 & 133.4 & 149.9 & 166.8 \\
& Set 2 & \% Correctly Classified & 67.5\% & 68.4\% & \textbf{65.2\%} & 66.3\% & 67.1\% & 67.7\% & 68.1\% & 68.2\% & 68.2\% & 68.2\% \\
 & & GMACs & 16.51 & 21.51 & \textbf{3.05} & 4.45 & 6.17 & 8.14 & 10.50 & 12.43 & 13.92 & 15.44 \\
& & Miliseconds & 67.4 & 143.5 & \textbf{35.6} & 44.7 & 55.1 & 66.6 & 80.2 & 91.4 & 99.9 & 108.7 \\
\hline
\end{tabular}}
\resizebox{0.80\textwidth}{!}{
\begin{tabular}{p{\linewidth}}
\vspace{-0.1 in}
\tiny Notes: See notes to Table \ref{tab:performance}. This table illustrates the performance on different datasets, sets of candidate models, and thresholds of the threshold-based classification model described in Figure \ref{fig:flowchart2}. For each dataset, the top benchmark shows the performance and complexity associated with the model that achieves the highest level of accuracy on the test sample. For the MNIST-style images, that model is LeNet-5. For SVHN and CIFAR-10, those models are DLA-Simple and ResNet-18, respectively. For the ImageNet and ImageNet-V2 datasets, it is ResNeXt-101-32x8d. The ``Max Propensity'' column shows the result of a propensity-based approach that ignores computational complexity and simply selects the classifier that has the highest confidence in its projection. For the Larger RGB images in panel C, the ``All'' rows consider an approach in which all 12 classifiers from the previous graphs and tables are considered in increasing order of computational complexity:(1) MobileNet v3 small, (2) MobileNet v3 large, (3) EfficientNet-b0, (4) AlexNet, (5) GoogLeNet, (6) Inception v3, (7) DenseNet-201, (8) EfficientNet-b7, (9) ResNet-101, (10) ResNeXt-101-32x8d, (11) VGG-19-bn, and (12) Wide ResNet 101-2, with GMACs per validation or test case ranging from 0.06 to 22.82. The ``Set 1'' and ``Set 2'' rows in the same panel present results for a restricted set of networks whose average performance relative to their computational intensity is relatively high. Set 1 includes (1') MobileNet v3 small, (2') EfficientNet-b0, (3') DenseNet-201, (4') ResNet-101, (5') ResNeXt-101-32x8d, and (6') Wide ResNet 101-2, selections 1, 3, 7, 9, 10, and 12 from the full set. Set 2 removes ResNet-101 and Wide ResNet 101-2 so that only four networks are considered. The \textbf{bolded} values in each row indicate, among the threshold-based classifiers presented here, the one that minimizes data use subject to the constraint that its performance is at least 95\% as large as the top benchmark's.
\end{tabular}}
\caption{Accuracy and Computation Cost of Sequential Model Complexity-Based Classifier by Dataset, Model, and Propensity Threshold} 
\label{tab:gmacs_threshold}
\end{table}

In panel A of Table \ref{tab:gmacs_threshold}, the two candidates are logistic regression and LeNet-5. As the bolded numbers illustrate, at the 70\% threshold, the MMACs per test case are reduced relative to the benchmark by 82.1\%, 47.6\%, and 72.2\% for the three datasets while classification accuracy declines by only 2.0\%, 4.6\%, and 0.5\%. At the higher thresholds, computational complexity remains substantially lower than in the benchmark, and classification accuracy comes within fractions of a percent of the benchmark for MNIST and KMNIST and exceeds it for Fashion MNIST.

In panel B of Table \ref{tab:gmacs_threshold}, the four candidates are logistic, LeNet-5, ResNet-18, and DLA-Simple. As the bolded values show, when the 80\% threshold is used, the computational cost is 87\% lower than the benchmark for SVHN and 13\% lower than the benchmark for CIFAR-10, while accuracy declines by only 4.7\% and 3.4\%, respectively. Thus, as with the use of downsampled images, the efficiency gains are modest for the CIFAR-10 dataset. Values at other thresholds illustrate the tradeoffs between computation and accuracy, and for the CIFAR-10, the estimators using the higher thresholds achieve rates of accuracy that exceed that of the benchmark model.

For the ImageNet and ImageNet-V2 datasets in panel C, three groupings of candidate models are considered for the threshold-based approach: ``All'' considers all 12 models in increasing order of their cost in GMACs, ``Set 1'' drops the 6 weakest models, and ``Set 2'' drops an additional 2 weaker models; the models are listed in the footnotes to the table. As the bolded values for the ``All'' rows show, when the threshold-based approach is applied on the full set of 12 neural networks, the 70\% threshold achieves the most computational savings given a less than 5\% reduction in accuracy, with a GMAC reduction of 74.6\% on ImageNet for only a 3.75\% accuracy decline. The savings on ImageNet-V2 are less substantial, involving a 23.7\% cost reduction for a 3.29\% reduction in accuracy. In both cases, the maximum propensity approach yields slight improvements in accuracy over the benchmark model, but with a substantial cost increase. The benefits of the threshold and max ensembles are more substantial when applied to the restricted set of networks, with the 50\% threshold yielding less than 5\% reduction in accuracy in all four cases considered here. For ImageNet, the reduction in computation costs are 85.4\% and 89.2\% from Set 1 and Set 2, with corresponding declines in accuracy of 2.46\% and 2.49\%. For ImageNet-V2, the savings are slightly smaller at 73.5\% and 81.5\% for Set 1 and Set 2, with declines in accuracy of 3.38\% and 3.41\%. For the maximum propensity approach, Set 1 more than doubles the computational cost but improves accuracy by 2.19\% for ImageNet and 2.80\% for ImageNet-V2. Set 2 increases computation cost by only 30.3\%, and it increases accuracy by 1.30\% and 1.32\% on the two datasets.

\begin{figure}
\centering
\begin{subfigure}{0.48\textwidth}
  \centering
  \includegraphics[width=\textwidth]{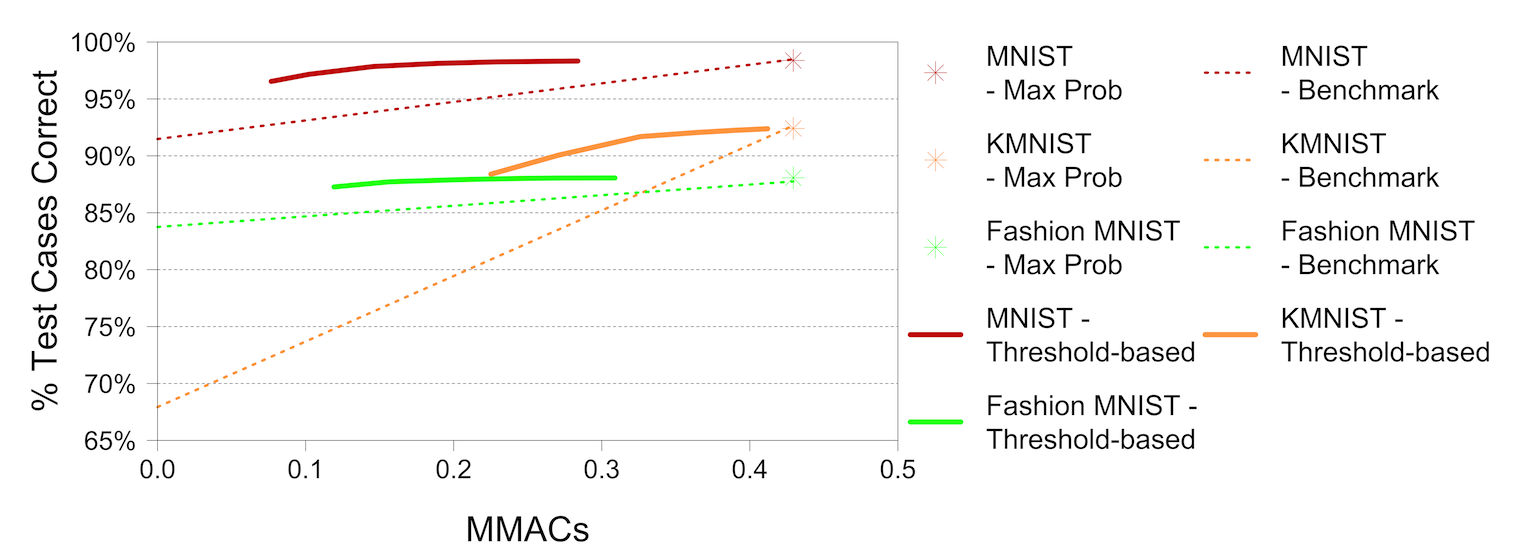}
  \caption{MNIST-style}
  \label{fig:mnist_ppf}
\end{subfigure}
\begin{subfigure}{0.48\textwidth}
  \centering
  \includegraphics[width=\textwidth]{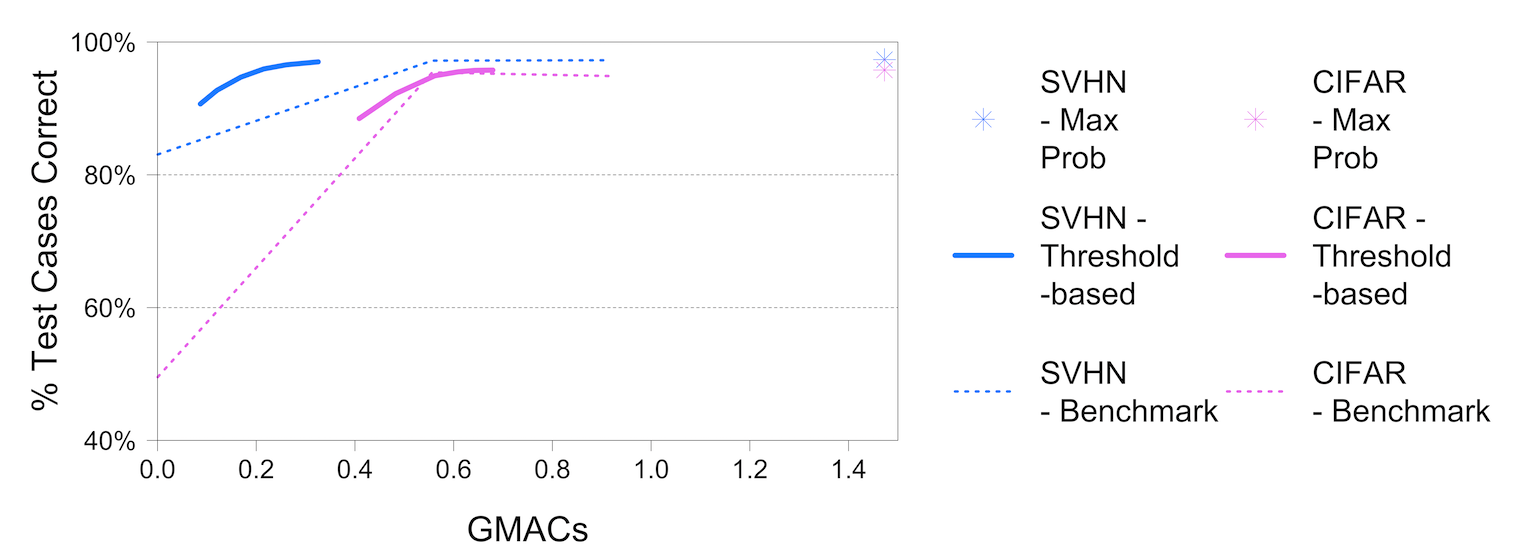}
  \caption{Small RGB}
  \label{fig:rgb_ppf}
\end{subfigure}
\begin{subfigure}{0.48\textwidth}
  \centering
  \includegraphics[width=\textwidth]{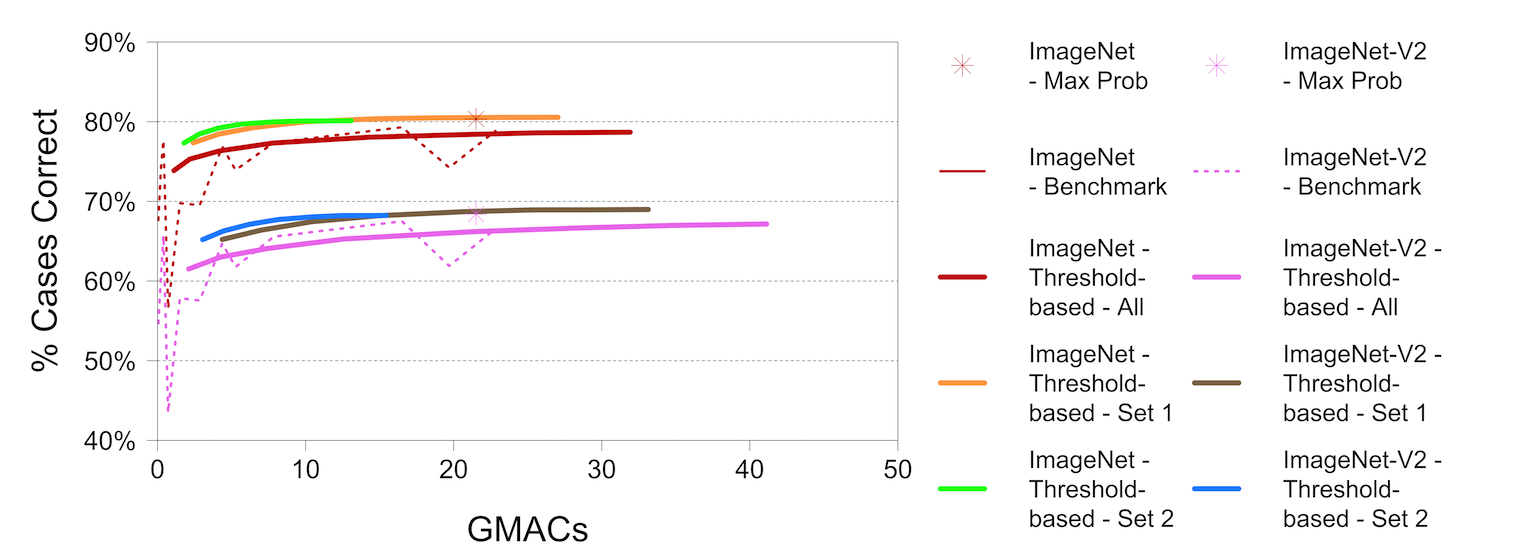}
  \caption{Larger RGB}
  \label{fig:inet_gmac_ppf}
\end{subfigure}
\begin{minipage}{\linewidth}
\vspace{0.10 in}
\flushleft \small Notes: See notes to Table \ref{tab:gmacs_threshold}. Symmetrically to Figure \ref{fig:ppf}, these results are the same as those presented in Table \ref{tab:threshold} but are presented graphically in order to make the tradeoff between performance and cost more apparent. The legend for Figures \ref{fig:ppf4} and \ref{fig:ppf5} is split across those two figures, and each of those two graphs shows each list benchmark and threshold-based results for all twelve of the neural networks considered for those datasets.
\end{minipage}
\caption{Accuracy versus Computation Time of Threshold-based Estimators}
\label{fig:gmacs_ppf}
\end{figure}

In order to better illustrate the tradeoffs between accuracy and computational complexity that the estimators in Table \ref{tab:gmacs_threshold} present, these relationships are illustrated graphically in Figure \ref{fig:gmacs_ppf}. The layout is similar to that of Figure \ref{fig:ppf} but with computational complexity rather than data usage plotted along the horizontal axis. The threshold-based approaches are illustrated by the solid lines. The values of accuracy and complexity that are shown here are the same values as appear in Table \ref{tab:gmacs_threshold}. The thin dashed lines in all four panels show the combinations of accuracy and cost that are achievable using the estimators on their own, with no thresholds; these lines illustrate the boundary of what is possible using the component neural networks without making use of the additional information of their levels of confidence in their projections. The efficiency gains produced by the threshold-based estimator are measured by the extent to which they expand this boundary upward (toward higher accuracy) and to the left (toward lower computational cost). The maximum propensity estimators are indicated by the stars shown on the graph.

For the MNIST-style images in Figure \ref{fig:mnist_ppf} and the small RGB images in Figure \ref{fig:rgb_ppf}, the threshold-based approach improves upon the benchmark to varying degrees. For MNIST, KMNIST, and SVHN shown in red, orange, and blue, the threshold-based estimators show marked improvements over the benchmarks. For Fashion MNIST, shown in green, the gains are less pronounced, and for CIFAR-10, the threshold-based approach hews close to the boundary illustrated by the benchmark. For the larger RGB images, the threshold-based approach that uses all 12 candidate networks performs comparably to the benchmark lines in both ImageNet and ImageNet-V2. One potential factor hindering the performance of the 12-network threshold-based approach is variation across networks in the reliability of the propensity scores that can be seen in the plots of accuracy versus self-assessed confidence in the left-hand panels of Figures \ref{fig:correct_imagenet} and \ref{fig:correct_imagenetv2}. In particular, Inception v3 tends to overstate its confidence, which may cause the ``All'' threshold-based approach to overuse projections from that classifier. When the restricted sets are used for the threshold-based approach, a consistent and moderate-sized expansion can be seen in the accuracy-cost boundary relative to the dashed benchmark lines.

\section{Discussion and Conclusion} \label{downsampling conclusion}

This study presents and evaluates a new ensemble-based approach for reducing the resource intensity of AI-based classification approaches. In the first of two applications, data usage is reduced by up to 85\% with less than 5\% loss in accuracy initially classifying low-resolution versions of images and only examining full-sized versions when the earlier pass lacks confidence in its projection. In the second application, computation time is reduced by up to 89\% with less than 5\% loss in accuracy by first using a simple model and only employing a complex model when the earlier pass lacks confidence. The method is versatile and does not require any additional training. The data and computation reductions vary in understandable ways and are concentrated among classifiers that use resources inefficiently.

One caveat to consider when applying this ensemble approach is its sensitivity to the design of the sequence of candidate classifiers. In the image resolution application, Table \ref{tab:performance} showed that the coarsest images were essentially unusable, and in the computation time application, Table \ref{tab:gmacs_threshold} showed that the threshold-based estimator performed best when weaker models were dropped from consideration to reduce needless resource use and the risk of incorrect classification. A related caveat is the comparability of the confidence measures across the candidate classifiers. For simplicity, no model-specific adjustments were applied to the propensity scores or thresholds, but if degrees of over- or under-confidence vary substantially across candidates, such adjustments might be appropriate.

One worthwhile future direction for this line of research is to test additional applications. The versatility of the design makes it well-suited for use in Natural Language Processing or speech recognition---two areas in which classification models vary widely in their resource-intensity. Another potential future direction for research involves customizing the candidate classifiers. This study's ensemble approach is designed for ease-of-use and does not require application-specific tuning or training like early exit approaches do. At the same time, however, the pre-trained candidate classifiers are not optimized for the pixelated data considered in the image resolution application. Potential enhancements to the model fitting process might augment the training data with combinations of images at different scales or resolutions as in \cite{Mishra2021} or \cite{Massoli2020} or with composite low/high resolution images in the vein of the Mix-up approach of \cite{Zhang2018}. Alternatively, each of the candidate classifiers could reply upon separate training sets or even separate model structures. While this study's approach substantially reduces data costs in many cases, a minor increase in the effort required for new applications may have the potential to reduce data costs even further by increasing performance for the low-cost candidates. Exploring application-specific training processes could help to identify such possibilities.

\printcredits

\bibliographystyle{IEEEtranN}

\bibliography{effort}

\vspace{-0.7 in}

\bio{}
\includegraphics[width=0.15\textwidth,angle=-90]{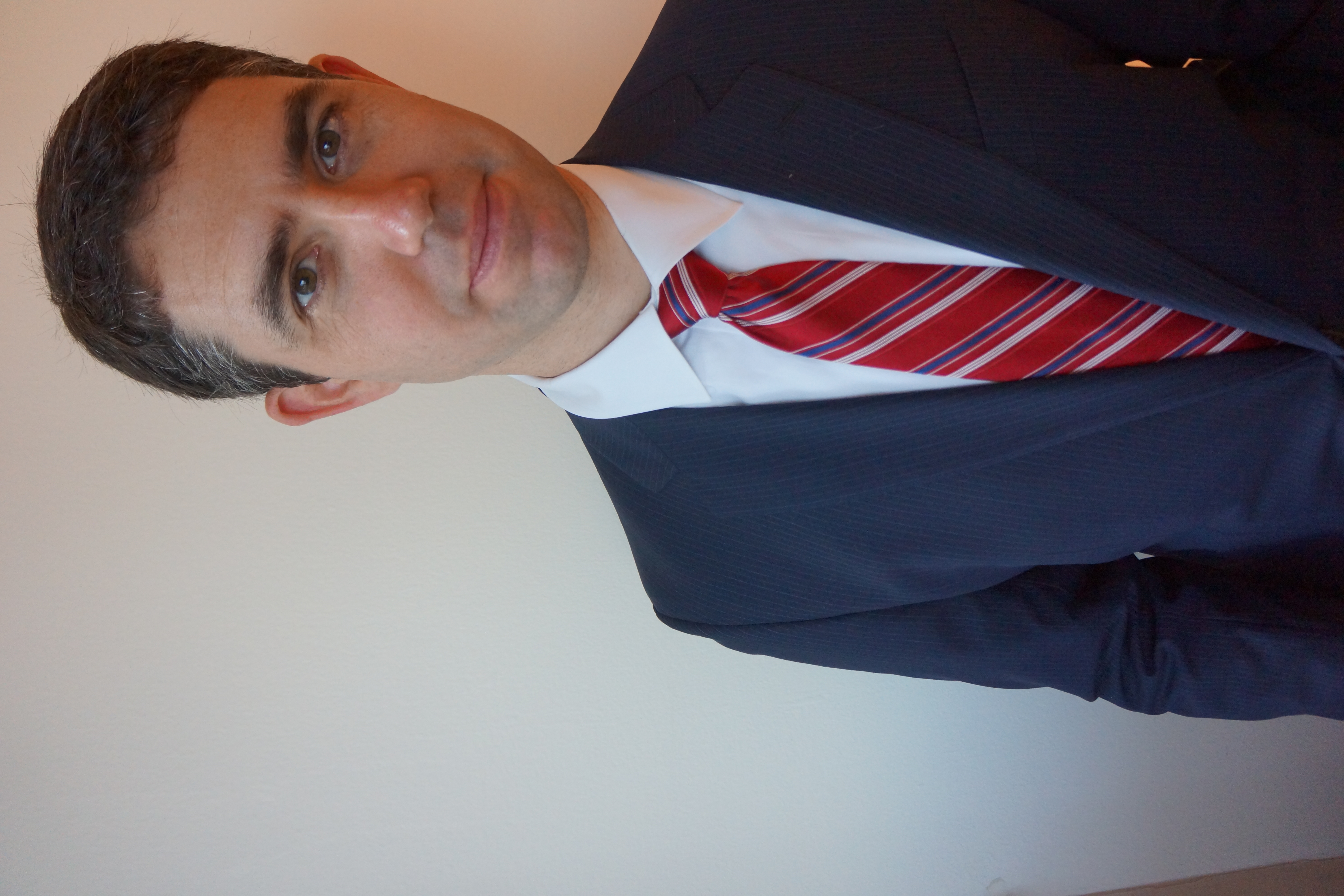}
\begin{minipage}{0.6\linewidth}
\flushleft 
\vspace{1 in}
Dr. Rohlfs is a Wall Street finance quant with doctorates in Economics and Electrical Engineering (expected); before joining the financial industry, he taught Economics and Syracuse University. His research examines artificial intelligence and generalization in neural networks, consumer demand, particularly for intangibles, recommendation models, and applied statistics and econometrics.
\end{minipage}

\endbio

\end{document}